\documentclass[sigplan,10pt,nonacm]{acmart}

\renewcommand\footnotetextcopyrightpermission[1]{}
\AtBeginDocument{%
  }

\usepackage{amsthm}
\usepackage{cleveref}

\crefname{observation}{observation}{observations}
\Crefname{observation}{Observation}{Observations}
\usepackage{enumitem}
\usepackage{array}
\usepackage{booktabs}
\usepackage{ulem}
\usepackage{makecell}
\usepackage{multirow}
\usepackage{adjustbox}
\acmConference[Conference acronym 'XX]{Make sure to enter the correct
  conference title from your rights confirmation emai}{June 03--05,
  2018}{Woodstock, NY}
\begin{document}
% \pagestyle{empty}
%%
%% The "title" command has an optional parameter,
%% allowing the author to define a "short title" to be used in page headers.
% \title{A Dependency-Aware Framework for Multi-GPU Temporal Graph Neural Networks Training}
\title{A Dependency-Aware Multi-GPU Training Framework for Temporal Graph Neural Networks}
%%
%% The "author" command and its associated commands are used to define
%% the authors and their affiliations.
%% Of note is the shared affiliation of the first two authors, and the
%% "authornote" and "authornotemark" commands
%% used to denote shared contribution to the research.
\author{Yang Zhao}
\authornote{Both authors contributed equally to this research.}
\email{yang_zhao@mail.ustc.edu.cn}
\affiliation{%
  \institution{University of Science and Technology of China}
  \city{Hefei}
  \state{Anhui}
  \country{China}
}

\author{Yue Dai}
\authornotemark[1]
\email{ydai21@illinoistech.edu}
\affiliation{%
  \institution{Illinois Institute of Technology}
  \city{Chicago}
  \state{Illinois}
  \country{USA}
}

\author{Liang Qiao}
\email{ql1an9@mail.ustc.edu.cn}
\affiliation{%
  \institution{University of Science and Technology of China}
  \city{Hefei}
  \state{Anhui}
  \country{China}
}

\author{Buxin Tu}
\email{tubuxin@mail.ustc.edu.cn}
\affiliation{%
  \institution{University of Science and Technology of China}
  \city{Hefei}
  \state{Anhui}
  \country{China}
}

\author{Bo Liu}
\email{liubo22@mail.ustc.edu.cn}
\affiliation{%
  \institution{University of Science and Technology of China}
  \city{Hefei}
  \state{Anhui}
  \country{China}
}

\author{Sen Zhang}
\email{sen@mail.ustc.edu.cn}
\affiliation{%
  \institution{University of Science and Technology of China}
  \city{Hefei}
  \state{Anhui}
  \country{China}
}

\author{Yi Zhang}
\email{yi_zhang@mail.ustc.edu.cn}
\affiliation{%
  \institution{University of Science and Technology of China}
  \city{Hefei}
  \state{Anhui}
  \country{China}
}

\author{Xiaoyu Hao}
\email{haoxiaoyu@mail.ustc.edu.cn}
\affiliation{%
  \institution{University of Science and Technology of China}
  \city{Hefei}
  \state{Anhui}
  \country{China}
}

\author{Jun Shi}
\email{shijun18@ustc.edu.cn}
\affiliation{%
  \institution{University of Science and Technology of China}
  \city{Hefei}
  \state{Anhui}
  \country{China}
}

\author{Junshi Chen}
\email{cjuns@ustc.edu.cn}
\affiliation{%
  \institution{University of Science and Technology of China}
  \city{Hefei}
  \state{Anhui}
  \country{China}
}

\author{Hong An}
\email{han@ustc.edu.cn}
\affiliation{%
  \institution{University of Science and Technology of China}
  \city{Hefei}
  \state{Anhui}
  \country{China}
}

%%
%% By default, the full list of authors will be used in the page
%% headers. Often, this list is too long, and will overlap
%% other information printed in the page headers. This command allows
%% the author to define a more concise list
%% of authors' names for this purpose.
\renewcommand{\shortauthors}{Trovato et al.}

%%
%% The abstract is a short summary of the work to be presented in the
%% article.
\begin{abstract}
Temporal graph neural networks (TGNNs) achieve superior performance in many dynamic graph learning tasks. These models use node state vectors (i.e., memories) to capture and record graph changes (i.e., events), and then exploit these memories to assist prediction.
However, this memory-centric design makes multi-GPU training challenging because predictions for later events often depend on memory updates from earlier events. 
To maintain model quality, existing multi-GPU TGNN training schemes strictly preserve all event dependencies, which limits parallelism and causes expensive cross-GPU memory synchronization.
In this work, we observe that many of these dependencies can be handled more flexibly without harming training quality.
Specifically, (a) dependencies in TGNN training have high locality in both time and graph topology, (b) the short-range dependencies missed within a large batch can often be recovered cheaply, and (c) many memory updates are not immediately needed. 
Based on these observations, we propose DepBloom to accelerate multi-GPU TGNN training with three techniques: (1) temporally bounded spatial partitioning to reduce unnecessary cross-GPU synchronization and improve parallelism; (2) intra-batch memory refinement to recover missing short-range temporal dependencies inside enlarged batches; and (3) critical-path optimization to prioritize urgent memory updates while deferring non-critical work during multi-GPU pipeline execution.
Experimental results across representative models and diverse datasets show that DepBloom achieves an average speedup of 2.84$\times$ and up to 5.16$\times$ over state-of-the-art methods, while maintaining comparable or better model accuracy.
\end{abstract}

%%
%% The code below is generated by the tool at http://dl.acm.org/ccs.cfm.
%% Please copy and paste the code instead of the example below.
%%
% \begin{CCSXML}
% <ccs2012>
%  <concept>
%   <concept_id>10010520.10010553.10010562</concept_id>
%   <concept_desc>Computer systems organization~Embedded systems</concept_desc>
%   <concept_significance>500</concept_significance>
%  </concept>
%  <concept>
%   <concept_id>10010520.10010575.10010755</concept_id>
%   <concept_desc>Computer systems organization~Redundancy</concept_desc>
%   <concept_significance>300</concept_significance>
%  </concept>
%  <concept>
%   <concept_id>10010520.10010553.10010554</concept_id>
%   <concept_desc>Computer systems organization~Robotics</concept_desc>
%   <concept_significance>100</concept_significance>
%  </concept>
%  <concept>
%   <concept_id>10003033.10003083.10003095</concept_id>
%   <concept_desc>Networks~Network reliability</concept_desc>
%   <concept_significance>100</concept_significance>
%  </concept>
% </ccs2012>
% \end{CCSXML}

% \ccsdesc[500]{Computer systems organization~Embedded systems}
% \ccsdesc[300]{Computer systems organization~Redundancy}
% \ccsdesc{Computer systems organization~Robotics}
% \ccsdesc[100]{Networks~Network reliability}

%%
%% Keywords. The author(s) should pick words that accurately describe
%% the work being presented. Separate the keywords with commas.
\keywords{Deep Learning, Dynamic Graph, Temporal Graph Neural Networks, Efficient Multi-GPU Training}
%% A "teaser" image appears between the author and affiliation
%% information and the body of the document, and typically spans the
%% page.
% \begin{teaserfigure}
%   \includegraphics[width=\textwidth]{sampleteaser}
%   \caption{Seattle Mariners at Spring Training, 2010.}
%   \Description{Enjoying the baseball game from the third-base
%   seats. Ichiro Suzuki preparing to bat.}
%   \label{fig:teaser}
% \end{teaserfigure}

% \received{20 February 2007}
% \received[revised]{12 March 2009}
% \received[accepted]{5 June 2009}

%%
%% This command processes the author and affiliation and title
%% information and builds the first part of the formatted document.
\newcommand{\M}{\textsc{xxx}}

\newcommand{\bred}[1]{\textcolor{red}{\sf\bfseries #1}}
\newcommand{\blue}[1]{\textcolor{blue}{#1}}
\newcommand{\yellow}[1]{\textcolor{yellow}{#1}}
\newcommand{\purple}[1]{\textcolor{purple}{#1}}
\newcommand{\red}[1]{\textcolor{red}{#1}}
\newcommand{\olive}[1]{\textcolor{olive}{#1}}
\newcommand{\brown}[1]{\textcolor{brown}{#1}}
\newcommand{\cross}[1]{\textcolor{red}{\sout{#1}}}

\newcommand{\Yue}[1]{\textcolor{teal}{\textbf{Yue: #1}}}
\newcommand{\Yang}[1]{\textcolor{olive}{\textbf{Yang: #1}}}
\newcommand{\Solved}[1]{}
\newcommand{\ToChecked}[1]{\yellow{\textbf{ToChecked:} #1}}
\newcommand{\Unsolved}[1]{\red{\textbf{Unsolved:} #1}}

\newcommand*\circled[1]{\tikz[baseline=(char.base)]{
  \node[shape=circle,draw,fill=black,text=white,font=\bf,inner sep=0.5pt] (char)
  {\small#1};
}}
\newcommand{\textbfit}[1]{\textbf{\textit{#1}}}
\newcommand{\textitbf}[1]{\textbf{\textit{#1}}}

\newcommand*\circledwhite[1]{\tikz[baseline=(char.base)]{
  \node[shape=circle,draw,fill=white,text=black,font=\bf,inner sep=0.5pt] (char)
  {\small#1};
}}

\newcommand{\squishlist}{
   \begin{list}{$\bullet$}
    { 
    \setlength{\itemsep}{0pt}      \setlength{\parsep}{0pt}
      \setlength{\topsep}{3pt}       \setlength{\partopsep}{0pt}
      \setlength{\listparindent}{-2pt}
      \setlength{\itemindent}{-5pt}
      \setlength{\leftmargin}{1em} \setlength{\labelwidth}{0em}
      \setlength{\labelsep}{0.5em} } }

\newcommand{\squishend}{
    \end{list}  }

\newcommand{\OurMethod}{DepBloom}

\newcommand{\eg}{\emph{e.g.}}

\newcommand{\zypre}[1]{}
\newcommand{\zyafter}[1]{\color{olive}#1\color{black}}

\newcommand{\yuepre}[1]{}
\newcommand{\yueafter}[1]{\color{teal}#1\color{black}}

\crefname{equation}{Equation}{Equation}
\crefname{figure}{Figure}{Figure}
\crefname{table}{Table}{Table}

% \newcommand{\testcase}[1]{\raisebox{-.2ex}{\protect\includegraphics[width=.7em,page=#1]{}}}

% \color{blue}
% TODO List:
% \begin{enumerate}
%     \item \color{red}{We need to shrink. Background to up to page-3, Motivation up to page-5, Design up to page-8.}
%     \item For 2.1, \Yue{We need to highlight the key feature (problem) of TGNN here: Given a message (a change on the graph), the update does not happen immediately but is triggered by a batch.}\Yang{Preliminarily Completed} \Yue{We may also need to make this consistent with our following figures. First, update (and clarify that the update relies on previous results), then embed.}
%     \item In the first paragraph, \sout{may consider more citations.} Can be shrunk if we need more space later. (Done)
%     \item \cref{tgnn} \sout{need to have a better caption. Since we have memory update before model prediction in our baseline, I would suggest to follow update->embedding->message generating step, or clarify that step 3 can also happened in before step 1 in the next batch.}
%     \item Add an algorithm for temporal split. (Hub-based time window split)
%     % \item 背景里面，observation 中容易提到了解决方法，可能会混淆。
%     \item \sout{Making all M-TGNN uniformly as TGNN (in background).}
%     \item \sout{Keep consistency on "Distributed" or "Multi-GPU". Maybe "Multi-GPU" is more suitable.}
%     \item \sout{font size should be enlarged as possible---use bolden Arial or Arial Narrow can help}
% \end{enumerate}
% \color{black}
% \newpage

\maketitle

% \Yang{TODO: maybe our framework is not distributed system. Instead it with "multi-GPU" M-TGNN is better.}
\section{Introduction} \label{sec: intro}

Dynamic graphs are widespread across various applications such as social networks~\cite{Kumar2018}, recommendation systems~\cite{song2023towards}, traffic forecasting~\cite{li2023dynamic}, and financial risk analysis~\cite{wushu2019}.
Compared with static graphs, dynamic graphs involve changing nodes and edges, which makes representation learning substantially more challenging~\cite{belli2020privacy,jia2018tequila,kumar2019predicting,rossi2020temporal}.
Temporal Graph Neural Networks (TGNNs) have emerged as an effective approach for learning on dynamic graphs and have achieved state-of-the-art performance on a range of dynamic graph tasks~\cite{cao2020spectral,longa2023graph,oskarsson2023temporal,rossi2020temporal,sankar2020dysat,wang2021apan,xu2020inductive,zhou2023disttgl,kazemi2020representation,trivedi2019dyrep,wang2024tglite,you2022roland,zhang2023tiger,dai2025memfreezing,dai2025towards,kumar2019predicting,emanuele2020temporal}.
TGNNs model a dynamic graph as a time-ordered sequence of \textit{events}, where each event records a graph change, such as adding a node or an edge.
A key feature of these models is that each node maintains a time-evolving state vector, often called \textit{node memory}~\cite{zhou2022tgl,gao2024etc,dai2025cascade,da2020distdgl,liu2025pipetgl,dai2025memfreezing}, which is updated when relevant events occur around it.
As dynamic graphs continue to grow in scale to billions or even trillions of events~\cite{leskovec2010signed,leskovec2016snap,zhou2022tgl}, efficient multi-GPU training of TGNNs is crucial for practical deployment.
% \Yue{Consider quickly discussing CTDG in this paragraph.}

However, TGNN training is fundamentally bottlenecked by its unique memory update mechanism.
Ideally, TGNNs follow a strict chronological execution order, updating node memories whenever they become relevant to an event~\cite{kumar2019predicting,emanuele2020temporal,wang2021apan,sankar2020dysat}. 
In TGNN training, however, events are grouped into \textit{batches}, where each batch reads memories from earlier batches for computing and writes back updated memories for later batches~\cite{kumar2019predicting,emanuele2020temporal,wang2021apan,sankar2020dysat,rossi2020temporal,zhou2022tgl,dai2025cascade,zhou2023disttgl,liu2025pipetgl}.
As a result, unlike conventional deep learning workloads, where batches are largely independent, TGNN batches are linked by strict cross-batch memory dependencies, leading to two major challenges in scaling multi-GPU TGNN training.
% \zyafter{
% \textbf{First, temporal dependency forces a trade-off between parallelism and training quality.} 
% Larger batches improve hardware utilization, but they delay memory updates and ignore the dependencies among events within the same batch. 
% While prior works~\cite{dai2025cascade,gao2024etc} improve single-GPU TGNN training by adaptively enlarging batches while controlling information loss, they are insufficient for multi-GPU setups: when batch sizes vary substantially, the workload across GPUs becomes uneven, leading to low GPU utilization. 
% Moreover, batching only provides limited parallelism under strict temporal dependencies, which is insufficient to saturate multi-GPU computing resources. 
\textbf{First, temporal dependency forces a trade-off between parallelism and training quality.} 
Larger batches improve hardware utilization, but they delay memory updates and ignore dependencies among events within the same batch. 
% Existing TGNN training systems designed for single-GPU execution, such as ETC and Cascade~\cite{dai2025cascade,gao2024etc}, enlarge batches while controlling information loss, but only explore partitioning along a single chronological event stream, leaving the potential of multiple parallel timelines unexplored.
% In multi-GPU settings, the resulting irregular batches are also difficult to distribute evenly across GPUs, often leading to load imbalance and pipeline bubbles.
Existing single-GPU TGNN systems, such as ETC and Cascade~\cite{dai2025cascade,gao2024etc}, improve efficiency by enlarging batches along a chronological event stream while controlling information loss. However, simply enlarging batches remains insufficient to expose scalable parallelism across GPUs, while the resulting irregular workloads can further cause load imbalance and pipeline bubbles in multi-GPU execution.
\textbf{Second, multi-GPU scalability is limited by expensive cross-GPU memory synchronization.}
% Existing systems~\cite{liu2025pipetgl,da2020distdgl} must synchronize updated memory states across GPUs at each dependency boundary before subsequent computation can proceed, introducing heavy communication overhead and prolonging pipeline stalls.
% , and limiting scalability.
% }
% \Yue{Comparison table OPT1.Existing not resolve problems.}
% \Yue{replace from "Existing system"}
% \color{teal}
Existing systems~\cite{liu2025pipetgl,da2020distdgl} either synchronize updated memory states across GPUs at the end of each batch or ignore such synchronization altogether. As a result, they either incur substantial communication overhead and prolonged pipeline stalls, or suffer noticeable training-quality degradation. 
Table~\ref{tab:method_compare} summarizes how representative prior methods, together with our design, address these two core challenges. 
As the table shows, prior methods partially improve one challenge or the other, but remain insufficient for scalable multi-GPU TGNN training.
% \color{black}

\begin{table}[t]
    \centering
    \small
    \setlength{\tabcolsep}{4pt}
    \caption{
    Comparison with representative efficient TGNN training systems under the two core challenges of scalable multi-GPU training.
    $\surd$ and $\blacktriangle$ denote explicit and partial support, respectively.
    }
    \begin{adjustbox}{max width=\columnwidth}
    \begin{tabular}{c|cc|cc}
    \hline
    \multirow{2}{*}{\textbf{Work}} 
    & \multicolumn{2}{c|}{\textbf{Quality-Aware Parallelism}}
    & \multicolumn{2}{c}{\textbf{Memory Sync Overhead}} \\
    \cline{2-5}
    & \textbf{Batching}
    & \textbf{Multi-GPU Scaling}
    & \textbf{Reduce Sync}
    & \textbf{Hide Sync} \\
    \hline
    ETC~\cite{gao2024etc} & $\surd$ & ~ & ~ & ~ \\
    Cascade~\cite{dai2025cascade} & $\surd$ & ~ & ~ & ~ \\
    PipeTGL~\cite{liu2025pipetgl} & ~ & $\surd$ & ~ & $\blacktriangle$ \\
    DistTGL~\cite{zhou2023disttgl} & ~ & $\blacktriangle$ & $\blacktriangle$ & ~ \\
    \hline
    DepBloom & $\surd$ & $\surd$ & $\surd$ & $\surd$ \\
    \hline
    \end{tabular}
    \end{adjustbox}
    \label{tab:method_compare}
\end{table}

Nevertheless, we find that many dependencies can be handled more flexibly without harming training quality. In particular, we make three observations:
First, \textbf{dependencies in TGNN training are often local in both time and graph topology}. 
% \zypre{
% This suggests a joint temporal-spatial partitioning strategy, where temporal windows preserve necessary synchronization across time, and topology-aware partitioning within each window exposes parallelism among weakly coupled subgraphs.
% }
% \zyafter{
% This locality, however, cannot be preserved well by partitioning along only one dimension: temporal-only partitioning breaks many topological dependencies, while spatial-only partitioning degrades training quality over long horizons.
This locality is hard to utilize with only one-dimensional partitioning. Temporal-only partitioning breaks topological dependencies while spatial-only partitioning degrades training quality over long horizons.
% }
\textbf{Second, the short-range temporal dependencies missed inside a large batch can often be recovered cheaply}.
Specifically, stale node memories in a large batch can be recovered based on earlier events within the same batch, enabling higher parallelism while maintaining training quality.
\textbf{Third, memory updates are not equally urgent for subsequent computation}. 
Only a small subset of updated node memories is immediately required by the next dependent batch. 
Therefore, uniformly synchronizing all updates places unnecessary work on the critical path and prolongs execution stalls.
Together, these observations suggest that scalable TGNN training could preserve only critical dependencies while relaxing or restructuring the rest to improve parallelism and reduce synchronization.
% \Yue{Comparison table OPT1.5. Existing not explore insights}

Based on these insights, we propose \OurMethod{}, an efficient framework for scalable multi-GPU TGNN training with three key techniques. First, \textit{temporally bounded spatial partitioning} divides the event stream into serial temporal windows and organizes each window into topology-aware parallel sub-chains, reducing unnecessary cross-GPU synchronization among sub-chains while preserving critical dependencies among windows. 
Second, \textit{intra-batch memory refinement} recovers missing short-range temporal information inside enlarged batches, enabling higher parallelism without sacrificing training quality. 
Third, \textit{critical-path optimization for multi-GPU pipeline execution} prioritizes urgent memory updates and removes non-critical updates and operations from the critical path, thereby improving pipeline execution efficiency and enhancing system scalability.
Experimental results across diverse benchmarks show that \OurMethod{} achieves an average speedup of 2.84$\times$ and up to 5.16$\times$ over baselines, while maintaining comparable or better model quality.
We summarize our contributions as follows:
\squishlist
    \item We identify the excessive preservation of event dependencies in existing multi-GPU TGNN training frameworks as a major limitation of parallelism and scalability, and recognize the potential of handling these dependencies more flexibly without harming training quality.
    \item We propose \OurMethod{}, a multi-GPU TGNN training framework that combines temporally bounded spatial partitioning, intra-batch memory refinement, and critical-path optimization to increase parallelism and reduce cross-GPU synchronization costs.
    \item We evaluate \OurMethod{} on representative TGNN models and dynamic graph datasets, and show that it substantially outperforms prior multi-GPU TGNN training systems while maintaining comparable or better accuracy.
\squishend

% \color{black}

% TODO: add our contribution.

\section{Background}
\subsection{Temporal GNN Learning on Dynamic Graphs}
\yuepre{
\textbf{Dynamic Graphs}
% 动态图用于描述节点和边随时间持续演化的系统，其中时间信息是定义图语义的重要组成部分。
Dynamic graphs describe systems whose vertices and edges evolve over time, where temporal information forms an essential part of the graph semantics.
% 常见的动态图表示方式包括离散时间动态图和连续时间动态图，前者由一系列快照构成，后者由按时间排序的事件流构成。
They are commonly represented as either discrete-time dynamic graphs (DTDGs), which are modeled as a sequence of graph snapshots, or continuous-time dynamic graphs (CTDGs), which are modeled as a chronologically ordered stream of timestamped events.
% 与 DTDG 相比，最近的研究越来越青睐 CTDG，因为它们保留了相互作用的精确时间，从而提供了更强的细粒度时间动态建模能力。
Compared with DTDGs, recent studies~\cite{rossi2020temporal,zhong2023dynamic,alilou2026temporal,hosseini2025deep} have increasingly favored CTDGs because they preserve the exact timing of interactions and thus provide a stronger capacity for modeling fine-grained temporal dynamics.
% 形式化地，连续时间动态图可表示为按时间排序的事件序列，其中每个事件对应一次带时间戳的节点交互或图更新。
Formally, a CTDG can be written as an ordered event sequence $\mathcal{E}=\{e(t_1), e(t_2), \ldots\}$ with $t_1 < t_2 < \cdots$, where each event $e(t_i)$ typically denotes a timestamped interaction or graph update occurring at time $t_i$.
% 在时刻$t_i$进行分析或预测时，只能依赖其之前已经发生的历史事件，这体现了动态图建模中的时间因果约束。
At time $t_i$, analysis or prediction is naturally conditioned on the historical prefix $\mathcal{E}_i^-=\{e(t_1), \ldots, e(t_{i-1})\}$, which reflects the temporal causality constraint inherent in dynamic graph modeling.
% 这种事件流视角也说明了动态图数据处理通常需要围绕历史依赖进行组织，包括时间约束的数据访问、邻域构建和状态维护等操作。
This event-stream view also implies that dynamic graph processing must be organized around historical dependencies, including time-respecting data access, neighborhood construction, and state maintenance.
% \Yue{No need to mention TGNN here; it is not introduced well yet, so it may be distracting.}\Yang{Done.}
\Solved{\Yue{We can make the notation consistent. For example, $\mathcal{E}=\{e(t_1), e(t_2), \ldots\}$ vs $\mathcal{E}=\{e_i\}_{i=1}^{T}$ in following sections: we may choose one of them}\Yang{OK, I choose $\mathcal{E}=\{e(t_1), e(t_2), \ldots\}$!}}
}

% \yueafter{
% \textbf{Dynamic Graphs}. Dynamic graphs model systems whose nodes and edges evolve over time. 
% They are commonly represented as either discrete-time dynamic graphs (DTDGs), modeled as graph snapshots, or continuous-time dynamic graphs (CTDGs), modeled as a chronologically ordered stream of timestamped events. 
% Recent work~\cite{rossi2020temporal,zhong2023dynamic,alilou2026temporal,hosseini2025deep} increasingly favors CTDGs because they preserve exact interaction timing and better capture fine-grained temporal dynamics. 
% Formally, a CTDG is an ordered event sequence $\mathcal{E}={e(t_1), e(t_2), \ldots}$ with $t_1 < t_2 < \cdots$, where each event denotes a timestamped graph interaction or update. 
% Prediction at time $t_i$ is naturally conditioned on the historical prefix $\mathcal{E}i^-={e(t_1), \ldots, e(t_{i-1})}$.
% \Yue{Discuss CTDG vs DTDG here.}

% \color{teal}
\noindent\textbf{Dynamic Graphs}. Dynamic graphs model systems whose nodes and edges evolve over time. 
They are commonly represented as either discrete-time dynamic graphs (DTDGs), modeled as a sequence of graph snapshots, or continuous-time dynamic graphs (CTDGs), modeled as a chronologically ordered stream of timestamped events. 
DTDGs can often be viewed as a coarser discretization of CTDG event streams, where events are grouped into time windows to form snapshots~\cite{dai2025cascade,dai2025memfreezing,zhou2022tgl}. 
Recent work~\cite{rossi2020temporal,zhong2023dynamic,alilou2026temporal,hosseini2025deep, zhou2022tgl, da2020distdgl,dai2025cascade,liu2025pipetgl} increasingly favors CTDGs because they preserve exact interaction timing and better capture fine-grained temporal dynamics. 
Formally, a CTDG is an ordered event sequence $\mathcal{E}={e(t_1), e(t_2), \ldots}$ with $t_1 < t_2 < \cdots$, where each event denotes a timestamped graph interaction or update. 
Prediction at time $t_i$ is based on the event history before $t_i$, which can be denoted as $\mathcal{E}_i^-={e(t_1), \ldots, e(t_{i-1})}$.
% \color{black}
% }
\begin{figure}[!t]
\centering
\includegraphics[width=\linewidth]{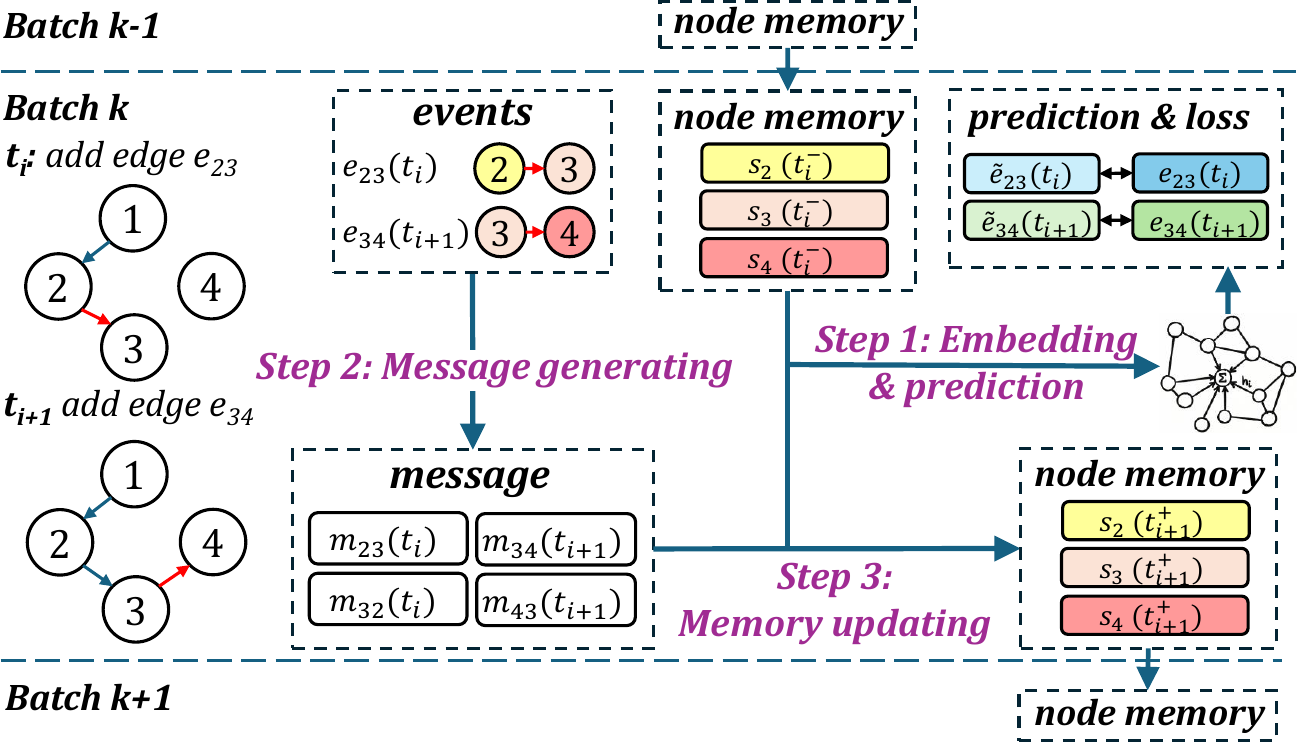}
\caption{
TGNN computing steps in one batch. In many TGNN training pipelines, the memory update of one batch is deferred and executed as the first step of the next batch.
}
\label{tgnn}
\end{figure}

\begin{figure*}[!t]
\centering
\includegraphics[width=\linewidth]{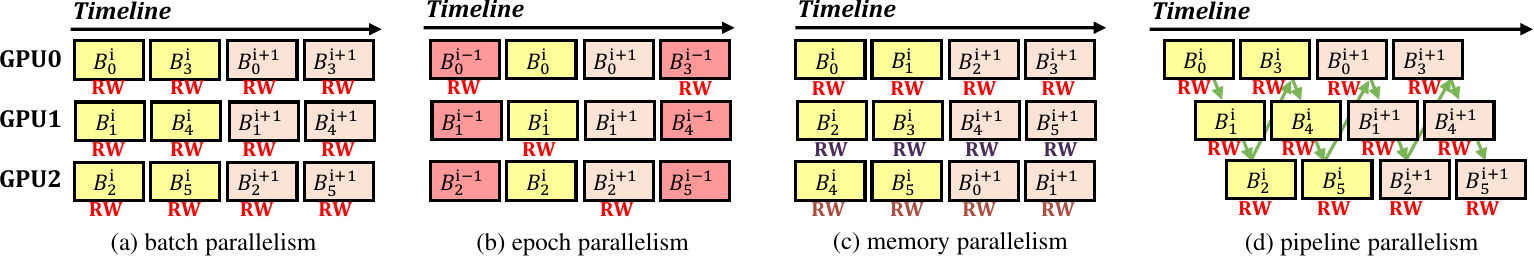}
\caption{
Comparison of four multi-GPU training paradigms for TGNNs: (a) batch parallelism, (b) epoch parallelism, (c) memory parallelism, and (d) pipeline parallelism. Here, $B_j^i$ denotes the $j$-th batch in epoch $i$, and different colors distinguish batches from different epochs. \textit{RW} denotes memory read and write operations, where different colors indicate different memory copies. Green arrows indicate inter-GPU data movement.}
% \ToChecked{\Yue{Would it be better to have communication paths than RWC to tell the dataflow?}\Yang{Add green arrows to indicate the dataflow between gpus.}}}
\label{stages}
\end{figure*}

\noindent\textbf{Temporal Graph Neural Networks}.
Temporal graph neural networks (TGNNs) maintain a time-evolving state vector for each node, commonly referred to as \textit{node memory}~\cite{rossi2020temporal,wang2021apan,zhou2022tgl,dai2025cascade,da2020distdgl,liu2025pipetgl}, to summarize its historical interactions for prediction. 
Formally, consider a chronologically ordered input event stream $\mathcal{E}=\{e_i\}_{i=1}^{T}$, where each event $e_i=(u_i,v_i,t_i,\mathbf{x}_{e_i})$ denotes an interaction between nodes $u_i$ and $v_i$ at time $t_i$, optionally with edge features $\mathbf{x}_{e_i}$. TGNNs update node memories and use them for predictions as follows:
When event $e_i$ arrives, the model first constructs messages using the pre-update memories of the two involved nodes, denoted by $\mathbf{s}_{u_i}^{-}$ and $\mathbf{s}_{v_i}^{-}$, together with the time gap $\Delta t_i$ and edge features $\mathbf{x}_{e_i}$. It then updates the corresponding node memories. For example, the memory update for node $u_i$ can be written as
\begin{equation}
\mathbf{m}_{v_i u_i}=\mathrm{msg}(\mathbf{s}^{-}_{v_i},\mathbf{s}^{-}_{u_i},\Delta t_i,\mathbf{x}_{e_i}),\quad
\mathbf{s}^{+}_{u_i}=\mathrm{UPDT}(\mathbf{s}^{-}_{u_i},\mathbf{m}_{v_i u_i}),
\label{eq:memory_update}
\end{equation}
and similarly for node $v_i$. Here, $\mathrm{msg}(\cdot)$ denotes message construction and $\mathrm{UPDT}(\cdot)$ denotes the memory update function.
For prediction, TGNNs typically compute time-aware node embeddings by feeding the target node memory and its temporal neighborhood memories into a GNN module, e.g.,
\begin{equation}
\mathbf{h}_u(t)=\mathrm{GNN}\ \!\Big(\mathbf{s}_u(t),\{\mathbf{s}_k(t)\mid k\in\mathcal{N}(u,t)\}\Big),
\label{eq:sgnn}
\end{equation}
followed by an MLP/scoring function to produce edge or node predictions.

\yuepre{
\noindent\textbf{Batched TGNN Training.}
% 在实际训练中，按时间排序的事件流会被划分为一系列连续 batch，并严格按照时间顺序依次处理。
In practice, a chronologically ordered event stream $\mathcal{E}=\{e_i\}_{i=1}^{T}$ is partitioned into a sequence of time-ordered batches $\{\mathcal{B}_k\}_{k=1}^{K}$ and processed strictly in temporal order.
% 如图所示，以边预测任务为例，一个 batch 的训练通常包含三个顺序执行的阶段：先进行 embedding 与 prediction，再生成 message，最后更新 node memory。
Using the edge prediction task as an example, \cref{tgnn} shows that the training of each batch $\mathcal{B}_k$ consists of three sequential stages: \emph{(1) embedding and prediction}, \emph{(2) message generating}, and \emph{(3) memory updating}.

% 第一步：读取 batch 开始前的 memory，并基于这些更新前状态完成表示计算与预测。
\textit{Step 1.} For the events in $\mathcal{B}_k$, the model first fetches the memories of the involved nodes before processing the current batch, denoted by $\mathbf{S}_k^-=\{\mathbf{s}_v(t^-)\}$, and uses these pre-update memories to compute node embeddings and make predictions.
% 以图中的事件为例，若 batch k 包含事件 e_{23}(t_i) 和 e_{34}(t_{i+1})，则它们在该阶段均基于节点 2、3、4 的更新前 memory 进行预测，而不会立即使用当前 batch 内新产生的 memory。
For instance, as shown in \cref{tgnn}, $\mathcal{B}_k$ contains events $e_{23}(t_i)$ and $e_{34}(t_{i+1})$, both events are predicted using the pre-update memories of nodes $2$, $3$, and $4$, rather than the newly generated memories within the same batch.

% 第二步：根据当前事件及其更新前 memory 生成 message，对应前文的 message 定义公式。
\textit{Step 2.} Messages are then generated from the current events and their pre-update memories according to \cref{eq:memory_update}, capturing the incremental effects of the batch on node states.

% 第三步：将 message 聚合并写回节点，对应前文的 memory update 公式。
\textit{Step 3.} These messages are aggregated and written back to the corresponding nodes following \cref{eq:sgnn}, transforming the pre-update memory $\mathbf{S}_k^-$ into the post-update memory $\mathbf{S}_k^+$. 

% 更新后的 memory 不会回流用于当前 batch 的预测，而是作为下一个 batch 的输入状态，以避免信息泄露。
The updated memory $\mathbf{S}_k^+$ is not reused for the prediction of the current batch; instead, it serves as the input state of the next batch to avoid information leakage.
% 因而，batch k 与 batch k+1 之间形成了严格的状态传递关系：batch k+1 必须读取由 batch k 产生的更新后 memory。
As a result, adjacent batches are connected by strict memory dependencies: batch $\mathcal{B}_{k+1}$ must consume the post-update memory produced by batch $\mathcal{B}_k$.
% 每个 batch 引起的 memory 更新会在 batch 之间形成严格依赖，从而给 TGNN 的并行扩展带来根本挑战。
Such batch-wise memory continuity fundamentally complicates the parallel scaling of TGNN training.
}

% \yueafter{
\noindent\textbf{Batched TGNN Training.}
In practice, the input event sequence $\mathcal{E}=\{e_i\}_{i=1}^{T}$ is divided into time-ordered batches $\{\mathcal{B}_k\}_{k=1}^{K}$ and processed sequentially. Using edge prediction as an example, \cref{tgnn} shows that each batch $\mathcal{B}_k$ has three stages:
(a) First, in the \textit{embedding and prediction stage}, the model fetches the pre-update memories of nodes involved in $\mathcal{B}_k$, denoted by $\mathbf{S}_k^-=\{\mathbf{s}_v(t^-)\}$, and uses them to compute embeddings and predictions. Thus, all events in the same batch are predicted from memories updated in previous batches.
(b) Next, in the \textit{message generation stage}, the model generates messages from the current events and their pre-update memories according to \cref{eq:memory_update}.
(c) Finally, in the \textit{memory update stage}, these messages are aggregated using a mean or most-recent operator~\cite{kumar2019predicting,rossi2020temporal}, and node memories are updated following \cref{eq:memory_update}. 

\subsection{Multi-GPU Training for TGNNs}
\yuepre{
% To enable parallel training of M-TGNNs on multiple GPUs, prior systems have explored four representative paradigms, as illustrated in \cref{stages}.
To enable multi-GPU TGNN training, prior work has explored four representative paradigms~\cite{zhou2023disttgl,liu2025pipetgl}, as illustrated in \cref{stages}. \textit{\textbf{Batch parallelism}}~\cite{zhou2023disttgl,zhong2023gnnflow,zhou2022tgl} assigns adjacent batches to different GPUs and executes them concurrently, improving throughput but relaxing inter-batch memory dependency, which leads to stale memories and degraded accuracy. 
\textit{\textbf{Epoch parallelism}}~\cite{zhou2023disttgl} overlaps different training epochs across GPUs to improve utilization. As a result, later epochs may process the same samples before updates from earlier epochs are fully incorporated, hurting training convergence. 
\textit{\textbf{Memory parallelism}}~\cite{zhou2023disttgl} assigns each GPU to train on a subsequence of the dynamic graph while maintaining multiple copies of node memory across GPUs, thereby reducing frequent synchronization on a single shared memory state. While this preserves local memory evolution within each worker, it increases memory overhead and departs from a globally unified memory update path.

% These paradigms represent different design choices for handling the tension among parallelism, memory continuity, and convergence behavior in multi-GPU M-TGNN training.
% \textit{\textbf{Batch parallelism}} improves throughput by executing adjacent batches concurrently on different GPUs, but it relaxes inter-batch memory continuity and therefore suffers from stale memories and degraded accuracy. \textit{\textbf{Epoch parallelism}} overlaps training across epochs to improve hardware utilization, yet repeated execution of the same batches under evolving parameters may increase gradient variance and hurt convergence. \textit{\textbf{Memory parallelism}} mitigates stale-memory effects by maintaining separate memory replicas or partitions across GPUs, but incurs substantial memory overhead and additional consistency cost. 

These three paradigms all improve hardware utilization by departing from the original unified memory evolution path of TGNN training. In contrast, \textit{\textbf{pipeline parallelism}}~\cite{liu2025pipetgl} preserves execution dependencies while still exposing concurrency. It decomposes the computation for each batch into dependent stages, such as node memory updates and node embeddings, and assigns one batch per GPU in a round-robin manner. Once a batch's node memory update is complete, the updated node memories are sent to the next GPU so it can begin processing the next batch, while the current GPU proceeds to the subsequent stage of the current batch. By overlapping multiple batches in this dependency-aware manner, pipeline parallelism improves GPU utilization without violating the original update order.
Among these paradigms, pipeline parallelism is therefore the most suitable for TGNN training\cite{liu2025pipetgl}, and we focus on it in this work.
}

% \yueafter{
To enable multi-GPU TGNN training, prior work has explored four representative paradigms~\cite{zhou2023disttgl,liu2025pipetgl}, as shown in \cref{stages}. \textbf{Batch parallelism}~\cite{zhou2023disttgl,zhong2023gnnflow,zhou2022tgl} executes adjacent batches concurrently on different GPUs, which improves throughput but causes stale memories by weakening inter-batch dependencies. \textbf{Epoch parallelism}~\cite{zhou2023disttgl} overlaps different training epochs across GPUs, but later epochs may process the same samples before earlier model updates are fully incorporated. \textbf{Memory parallelism}~\cite{zhou2023disttgl} reduces synchronization by keeping multiple copies of node memory across GPUs, but this increases memory overhead and disrupts memory evolution across the entire training event stream. These three paradigms improve hardware utilization by relaxing the original dependency order of TGNN training. 

In contrast, \textbf{pipeline parallelism}~\cite{liu2025pipetgl} preserves dependency order while still exposing concurrency. It decomposes each batch into dependent stages, such as node memory update and node embedding, and overlaps different batches across GPUs. Once one batch finishes its memory update, the next GPU can start the following batch, while the current GPU continues the later stages of the current batch. Because it improves concurrency without violating the original update order, pipeline parallelism is the most suitable paradigm for TGNN training~\cite{liu2025pipetgl}, and we focus on it in this work.
\Solved{
\Yue{(Q1) We did not define stages here...(Q2) Simply explaining pipeline parallelism is better seems weak...does pipeTGL quantitatively discuss why? If so, we can put a citation here, and a pointer to our evaluation results on DistTGL.}\Yang{Here, I add the citation of pipeTGL to quantitatively show the performance of pipeline parallelism}}

% \Yue{We may need a figure to illustrate these baselines. If pipeline parallelism wins in distributed TGNN training, we need to highlight why it wins and maybe shrink others.}\Yang{Done.}

\Solved{
\color{teal}
To enable multi-GPU TGNN training, prior work has explored four representative paradigms~\cite{zhou2023disttgl,liu2025pipetgl}, as illustrated in \cref{stages}. 

\textit{\textbf{Batch parallelism}}~\cite{zhou2023disttgl} assigns adjacent batches to different GPUs and executes them concurrently, improving throughput but relaxing inter-batch memory dependency, which leads to stale memories and degraded accuracy. 
\textit{\textbf{Epoch parallelism}}~\cite{zhou2023disttgl} overlaps different training epochs across GPUs to improve utilization. As a result, later epochs may process the same samples before updates from earlier epochs are fully incorporated, hurting training convergence. 
\textit{\textbf{Memory parallelism}}~\cite{zhou2023disttgl} assigns each GPU to train on a subsequence of the dynamic graph while maintaining multiple copies of node memory across GPUs, thereby reducing frequent synchronization on a single shared memory state. While this preserves local memory evolution within each worker, it increases memory overhead and departs from a globally unified memory update path.

These three paradigms all improve hardware utilization by departing from the original unified memory evolution path of TGNN training. In contrast, \textit{\textbf{pipeline parallelism}}~\cite{liu2025pipetgl} preserves execution dependencies while still exposing concurrency. It decomposes the computation for each batch into dependent stages, such as memory updates and node embeddings, and assigns one batch per GPU in a round-robin manner. Once a batch's memory update is complete, the updated memory is sent to the next GPU so it can begin processing the next batch, while the current GPU proceeds to the subsequent stage of the current batch. By overlapping multiple batches in this dependency-aware manner, pipeline parallelism improves GPU utilization without violating the original update order.
Among these paradigms, pipeline parallelism is therefore the most suitable for TGNN training, and we focus on it in this work.

\color{black}
}
% PipeTGL is a representative system following this design and further improves execution efficiency through dependency-aware scheduling and communication optimization. As a result, pipeline parallelism provides a practical basis for scalable multi-GPU M-TGNN training, although its performance can still be limited by communication overheads and pipeline bubbles.

% \input{src/3-motivation}
\section{Motivation and Opportunity}

\subsection{Bottlenecks in Multi-GPU TGNN Training}

Although recent pipeline-based systems can preserve temporal dependencies~\cite{liu2025pipetgl}, they still scale poorly. To quantify this limitation, we profile PipeTGL~\cite{liu2025pipetgl}, a state-of-the-art multi-GPU TGNN training system, on \texttt{WIKI-TALK}~\cite{leskovec2010signed} using 2 to 8 NVIDIA RTX 4090 GPUs.
As shown in \cref{fig:gpu_scaling}, adding more GPUs does not consistently reduce end-to-end latency. Although total latency decreases from 2 to 6 GPUs, the scaling quickly saturates, and 8 GPUs even underperform 6 GPUs. The overall speedup over 2 GPUs is only 1.77$\times$ with 6 GPUs and 1.46$\times$ with 8 GPUs. 
This poor scalability mainly stems from two fundamental challenges:

\begin{figure}[htbp]
\centering
\includegraphics[width=\linewidth]{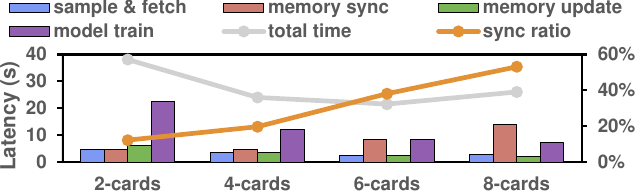}
\caption{Latencies of training TGN on \texttt{WIKI-TALK} with different numbers of GPUs. (1) \texttt{Sample\&fetch} includes neighbor sampling and input transfer from CPU to GPU; (2) \texttt{memory\_sync} denotes the transfer of updated node memories across GPUs; (3) \texttt{memory\_update} computes new node-memory states required by subsequent GPUs; and (4) \texttt{model\_train} includes forward propagation, backward propagation, and parameter updates.}
% \ToChecked{\Yue{We need to explain the stages. Also, I think we can keep either this or Figure 5---vote for this.}\Yang{I think figure 3 is better and add the ratio of sync on this.}}}
\label{fig:gpu_scaling}
\end{figure}

\begin{figure}[htbp]
\centering
\includegraphics[width=\linewidth]{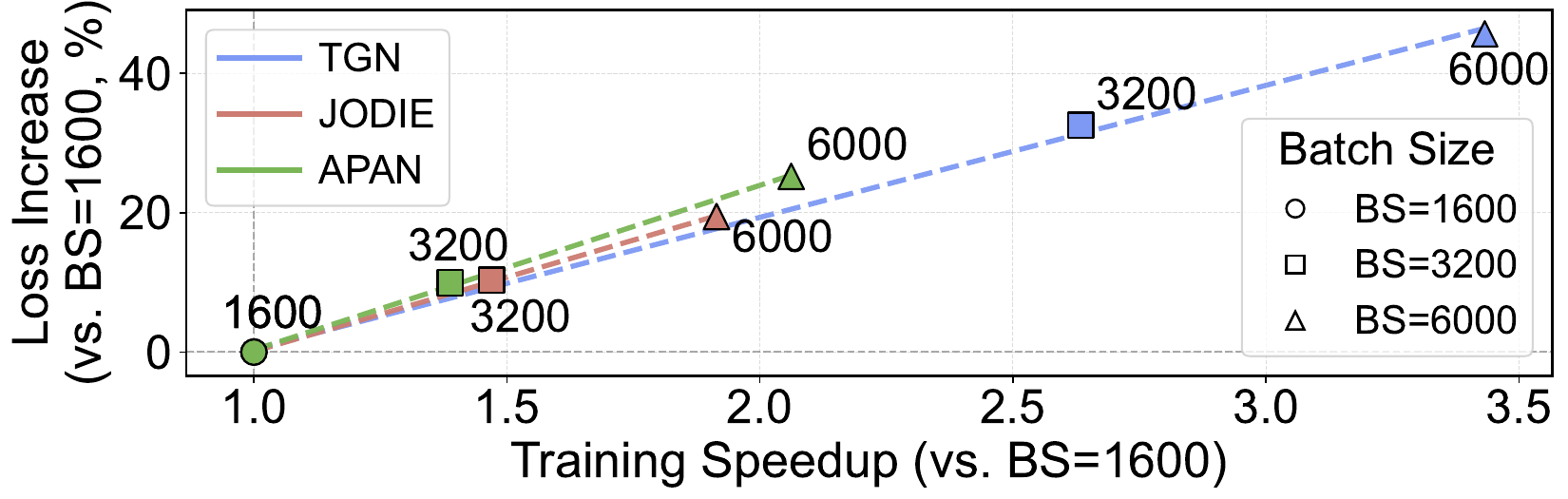}
\caption{Trade-off between training efficiency and model quality under different batch sizes on \texttt{WIKI-TALK}.}
\label{fig:batch_tradeoff}
\end{figure}

% \zypre{
% \textbf{\textit{Challenge 1: Temporal dependency forces a trade-off between parallelism and training quality.}}
% In TGNN training, memory updates generated within a batch are deferred until the batch finishes. 
% Consequently, later events in the same batch cannot benefit from updates produced by earlier related events, so larger batches improve parallelism but weaken short-range temporal dependency modeling and often hurt accuracy.
% As shown in \cref{fig:batch_tradeoff}, when training representative TGNNs including TGN~\cite{rossi2020temporal}, JODIE~\cite{kumar2019predicting}, and APAN~\cite{wang2021apan} on \texttt{WIKI-TALK} \cite{leskovec2010signed} with 8 GPUs, increasing the batch size from 1600 to 6000 consistently improves training throughput but degrades training quality. 
% The larger batch yields speedups of up to 3.84$\times$, 1.91$\times$, and 2.06$\times$, while increasing training loss by 45.7\%, 19.5\%, and 25.2\%, respectively. 
% As a result, existing multi-GPU TGNN training frameworks~\cite{liu2025pipetgl,zhou2023disttgl} use relatively small batches to preserve quality, but this also increases synchronization frequency and causes more frequent stalls on memory updates.
% }
% \zyafter{
\textbf{Challenge 1: Temporal dependency forces a trade-off between parallelism and training quality.}
In TGNN training, memory updates generated within a batch are deferred until the batch finishes. 
As a result, later events in the same batch cannot benefit from updates produced by earlier related events, so larger batches improve parallelism but weaken short-range temporal dependency modeling and often hurt accuracy.
As shown in \cref{fig:batch_tradeoff}, increasing the batch size consistently improves training throughput on \texttt{WIKI-TALK}~\cite{leskovec2010signed}, while causing clear loss degradation.
% Adaptive batching methods such as \textsc{Cascade} and ETC~\cite{dai2025cascade,gao2024etc} mitigate this trade-off by dynamically enlarging batches, but are primarily designed for single-GPU execution.
% \color{olive}
Such batch-centric designs are effective for single-GPU execution, but expose parallelism only along a single global event stream: once a dependency reaches the batch boundary, unrelated graph regions cannot progress independently.
Meanwhile, irregular dependency-aware batch sizes can cause workload imbalance and bubbles across GPUs, limiting multi-GPU scalability.
% Consequently, existing multi-GPU TGNN systems still rely on relatively small batches to preserve quality~\cite{liu2025pipetgl,zhou2023disttgl}, leading to frequent synchronization and stalls on memory updates.
% \color{black}
% In multi-GPU settings, substantial batch-size variation can further cause workload imbalance and reduce GPU utilization, while batching alone still exposes only limited parallelism under strict temporal dependencies. 
Consequently, existing multi-GPU TGNN systems still rely on relatively small batches to preserve quality~\cite{liu2025pipetgl,zhou2023disttgl}, which in turn increases synchronization frequency and leads to frequent stalls on memory updates.

% \zypre{
% \textbf{\textit{Challenge 2: Cross-GPU memory synchronization is expensive and scales poorly.}}
% }
% \zyafter{
\textbf{Challenge 2: Multi-GPU scalability is limited by expensive cross-GPU memory synchronization.}
% }
The second challenge is that cross-GPU memory synchronization is both expensive and increasingly inefficient at scale. 
In multi-GPU TGNN training, a batch cannot proceed until the required memory update has been produced by the previous GPU and transferred to the current one.
As a result, each synchronization point incurs not only waiting time but also explicit communication overhead. 
As the number of GPUs increases, this overhead becomes harder to hide and increasingly dominates execution.
This effect is clearly reflected in \cref{fig:gpu_scaling}. Although the computation stages continue to shrink as more GPUs are added, the synchronization latency rises from 4.70\,s on 2 GPUs to 13.85\,s on 8 GPUs, and becomes 2.92$\times$ higher than that on 4 GPUs.
Meanwhile, its share of total runtime increases sharply from 12.3\% to 53.1\%. 
These results show that cross-GPU memory synchronization becomes a dominant bottleneck as the system scales.

In summary, multi-GPU TGNN training scales poorly because (1) execution on each GPU repeatedly stalls due to memory updates across small batches, and (2) those stalls are too expensive to be effectively hidden by pipelining. This naturally raises the following question: can we reduce (i) the frequency of memory updates and (ii) the cost of each memory synchronization without hurting training quality?
% \color{black}
\Solved{
\Yang{In this challenge, I add the evidence of why sync can cause a few bubbles (about pipeline depth).} \Yue{Could be good to have these latency models. But maybe we can simplify this into a single statement if we do not use these equations for a cost model — currently, it requires imagining the pipeline workflow, which could be distracting. Only keep if you want to clarify that "sync cannot be hide at each batch so overall will be accumulated".}
\Yang{I think it's right. To understand this $T_{bubble}$, the reader must imagine the pipeline workflow and try to confirm why the deeper pipeline causes more bubbles. Maybe it really makes the reader distracted.}}

\subsection{Rethinking Dependencies in Multi-GPU TGNN Training}\label{subsec: motivation}

By revisiting dependency handling in TGNN training, we find that current execution strategies are often overly conservative, leaving substantial room to improve parallelism and reduce overhead. Specifically, we make three key observations that motivate our design.

Our first observation is that \textit{\uline{dependencies in TGNN training are often local in both time and graph topology, so effective partitioning must jointly consider both.}} Partitioning is a natural way to reduce cross-GPU communication~\cite{shao2024distributed,yang2023betty,gandhi2021p3}: 
% \zypre{if a training sequence can be decomposed into multiple largely independent units, which we call \textit{sub-chains}, then these units can execute in parallel while preserving sequential dependencies within each sub-chain and reducing synchronization across them. }
% \zyafter{
If a training sequence can be decomposed into multiple largely independent units, which we call \textit{sub-chains}, these units can execute in parallel. This preserves sequential dependencies within each sub-chain while reducing synchronization across sub-chains.
% }
For TGNN training, however, exploiting this opportunity requires partitioning that respects both temporal locality and graph topology.

\begin{figure}[htbp]
\centering
\includegraphics[width=\linewidth]{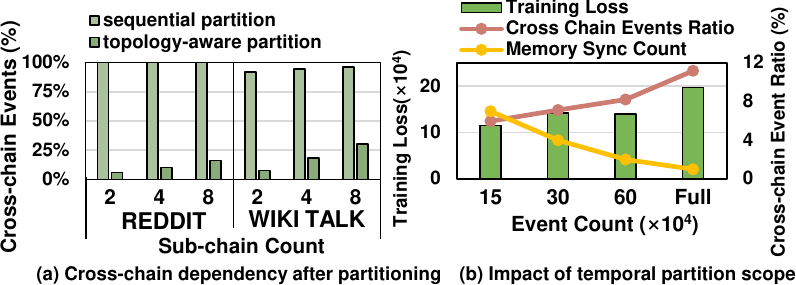}
\caption{(a) Cross-chain event ratio versus number of sub-chains, comparing naive sequential partitioning against topology-aware partitioning. 
% The original sequences span first 150,000 events on \texttt{REDDIT} and 300,000 events on \texttt{WIKI\_TALK}.
(b) Cross-chain event ratio and training loss as a function of temporal window size.
% \zypre{Ranging from 150,000 batches to the whole graph.}
% \zyafter{Delete something Here.}
}
\label{fig:motivation-chain}
\end{figure}

\textit{Limits of temporal partitioning.}
Existing approaches, such as memory parallelism~\cite{zhou2023disttgl}, partition the input sequence only along the temporal dimension and ignore graph topology. As a result, strongly coupled events, which share the same source nodes, are often assigned to different sub-chains, which breaks dependencies across sub-chains. To quantify this effect, we apply (i) sequential partitioning by dividing events chronologically~\cite{zhou2023disttgl} and (ii) topology-aware partitioning using a minimum-cut solver~\cite{karypis1997metis} on the first 150,000 events on \texttt{REDDIT} and 300,000 events on \texttt{WIKI\_TALK}, and then measure the fraction of dependencies broken across sub-chains. 
As shown in \cref{fig:motivation-chain}(a), on \texttt{REDDIT}, sequential partitioning yields \textbf{99.90\%}--\textbf{99.97\%} broken dependencies, whereas topology-aware partitioning reduces this ratio to \textbf{5.98\%}--\textbf{16.24\%}. On \texttt{WIKI\_TALK}, the ratio drops from \textbf{91.83\%}--\textbf{96.23\%} to \textbf{7.86\%}--\textbf{30.50\%}. The results show that temporal-only partitioning incurs substantial dependency breaks.

\textit{Limits of spatial partitioning.}
Directly applying spatial partitioning to the full dynamic graph, as is often done for static GNNs~\cite{zheng2022distributed,da2020distdgl,cai2021dgcl}, is too aggressive for TGNN training because it ignores temporal locality and allows omitted cross-chain dependencies to accumulate over long horizons. To evaluate this effect, we use the same setup as in \cref{fig:motivation-chain}(a), vary the temporal window length before partitioning, and train TGN~\cite{rossi2020temporal} on the resulting partitioned graphs. As shown in \cref{fig:motivation-chain}(b), when we directly partition the whole graph, the cross-chain event ratio increases from \textbf{5.98\%} to \textbf{11.19\%}, and training loss rises by \textbf{70.7\%}. These results show that spatial-only partitioning over the entire input dynamic graph may significantly degrade training quality.

% Together, these results show that effective TGNN partitioning must jointly consider temporal dependency and graph topology.

\begin{figure}[htbp]
\centering
\includegraphics[width=\linewidth]{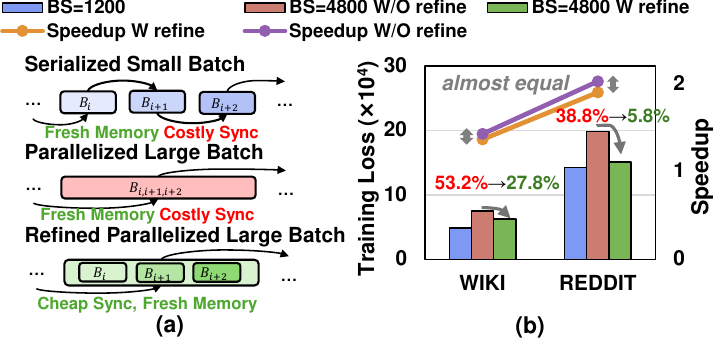}
\caption{
% Comparison of three TGNN batch execution strategies.
(a) Execution modes under different parallel strategies.
(b) Speedup versus training loss increase under larger batch sizes. Arrows show the percentage loss increase from BS=1200 to BS=4800.
}
\label{fig:motivation-refinement}
\end{figure}

Our second observation is that \textit{\uline{lost temporal dependencies in large batches can often be recovered cheaply.}} As shown in \cref{fig:motivation-refinement}(a), larger batches often hurt model quality because node memories miss short-range temporal dependencies among events within the same batch. However, much of this lost signal can be recovered through \textit{memory refinement}. 
Instead of executing several small batches sequentially, we split a large batch into consecutive micro-batches and let later micro-batches refine their input node memories using (i) the node memories carried over from the previous batch along with (ii) the \emph{most recent messages} from earlier micro-batches.
The refinement is performed by a lightweight RNN-based module that is trained jointly with the TGNNs.
More details are introduced in Section~\ref{subsec: design_mem_refine}. 
Since these messages can be directly accessed without waiting for intermediate memory updates, this enables parallel micro-batch execution without synchronization.

To quantify its potential, we use PipeTGL~\cite{liu2025pipetgl} to train TGN~\cite{rossi2020temporal} on \texttt{WIKI} and \texttt{REDDIT}~\cite{kumar2019predicting} with (a) default small batches (\texttt{Batch\_Size=1200}), (b) larger batches (\texttt{Batch\_Size =4800}) without refinement, and (c) larger batches with \textit{memory refinement}, which uses events within the same batch to update node memories in parallel (as detailed in Section~\ref{subsec: design_mem_refine}). As shown in \cref{fig:motivation-refinement}(b), compared with large-batch execution without refinement, 
% \textit{memory refinement} reduces the loss increase by 25.4\% on \texttt{WIKI} and 33.0\% on \texttt{REDDIT}, 
\textit{memory refinement} reduces the loss increase from \textbf{53.2\% to 27.8\%} on \texttt{WIKI} and from \textbf{38.8\% to 5.8\%} on \texttt{REDDIT},
while still achieving 1.4$\times$ and 1.9$\times$ speedup, respectively. 
These results show that much of the lost short-range temporal signal in large batches can be recovered without exact sequential memory updates.

Our third observation is that \textit{\uline{node memory updates are not equally urgent for subsequent computation,}} yet existing training schemes still synchronize them as a whole. In practice, only a small subset of updated nodes is immediately needed by the next batch, while most updated nodes are either consumed later or not revisited in the near term.

\begin{figure}[htbp]
\centering
\includegraphics[width=\linewidth]{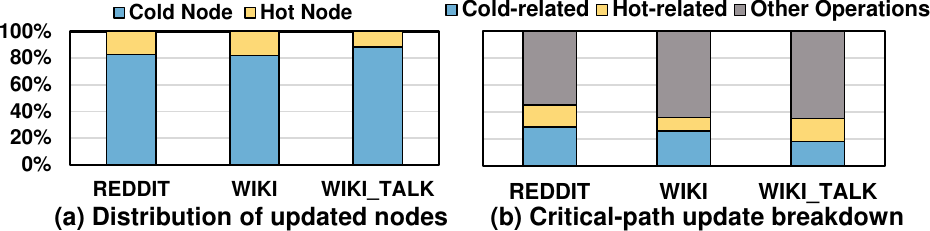}
\caption{Urgency mismatch in batch-produced memory updates. (a) Distribution of updated nodes immediately required by the next batch (\textit{Hot Node}) and those that are not (\textit{Cold Node}). (b) Breakdown of the critical path latency.
}
% \ToChecked{\Yue{Better to have a vertical line to devide results of REDDIT and WIKI\_TALK}} (b) Breakdown of latency in the critical update stage.\Solved{\Yue{latency? or operation counts?}}}
\label{fig:hot-cold}
\end{figure}

As shown in \cref{fig:hot-cold}(a), while training TGNN models on \texttt{WIKI}, \texttt{REDDIT}, and \texttt{WIKI\_TALK}, only \textbf{17.39\%}, \textbf{18.02\%}, and \textbf{11.57\%} of updated nodes, respectively, are immediately needed by the next batch; most are non-urgent. 
However, \cref{fig:hot-cold}(b) shows that these non-urgent updates still consume a substantial portion of stalled update time: \textbf{28.93\%}, \textbf{25.87\%}, and \textbf{18.21\%}, respectively. This unnecessary overhead is comparable to, or even larger than, that of immediately required updates (\textbf{16.18\%}, \textbf{10.15\%}, and \textbf{17.05\%}). This indicates that a significant fraction of synchronization-critical time is spent on updates that need not be consumed immediately.
These results show that current synchronization policies are overly conservative: they place non-urgent updates on the critical path, although the next batch does not immediately need them. 
Therefore, prioritizing urgent updates and deferring the rest can shorten stalls and reduce synchronization cost.

\section{Design}

\subsection{Overview}

\begin{figure}[htbp]
\centering
\includegraphics[width=\linewidth]{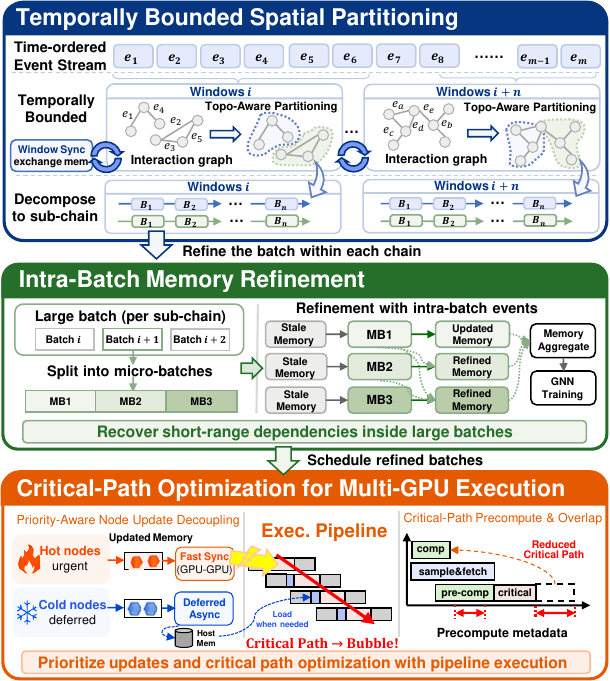}
\caption{Design overview of \OurMethod{}.}
\label{fig:overview}
\end{figure}

% \color{teal}
Based on the above observations, we propose \OurMethod{}, an efficient framework for multi-GPU TGNN training. \OurMethod{} integrates three complementary techniques to improve scalability and efficiency.
As illustrated in \cref{fig:overview}, \OurMethod{} improves multi-GPU TGNN training from top to bottom. 
First, \textbf{Temporally Bounded Spatial Partitioning} divides the training event sequence into serial temporal windows and further partitions each window into topology-aware sub-chains, exposing parallelism while minimizing cross-chain synchronization. 
Next, \textbf{Intra-Batch Memory Refinement} enables efficient large-batch execution within each sub-chain by compensating for missing short-range temporal dependencies caused by stale updates, thereby increasing parallelism without sacrificing training quality. 
Finally, \textbf{Critical-Path Optimization for Multi-GPU Pipeline Execution} prioritizes urgent memory updates and removes deferrable operations from the critical path, thereby reducing synchronization cost and shortening execution stalls.

% As illustrated in \cref{fig:overview}, \OurMethod{} improves multi-GPU TGNN training from the top to the bottom. \textit{Temporally Bounded Spatial Partitioning} first organizes the event stream into weakly coupled sub-chains within bounded temporal windows, reducing cross-chain dependencies and synchronization.
% Within each sub-chain, \textit{Intra-Batch Memory Refinement} enables larger-batch execution while recovering the lost short-range temporal signal. 
% Finally, \textit{Critical-Path Optimization for Multi-GPU Pipeline Execution} further improves efficiency by shortening the synchronization-critical path and mitigating pipeline stalls.
% \color{black}

\begin{figure*}[t]
\centering
\includegraphics[width=\linewidth]{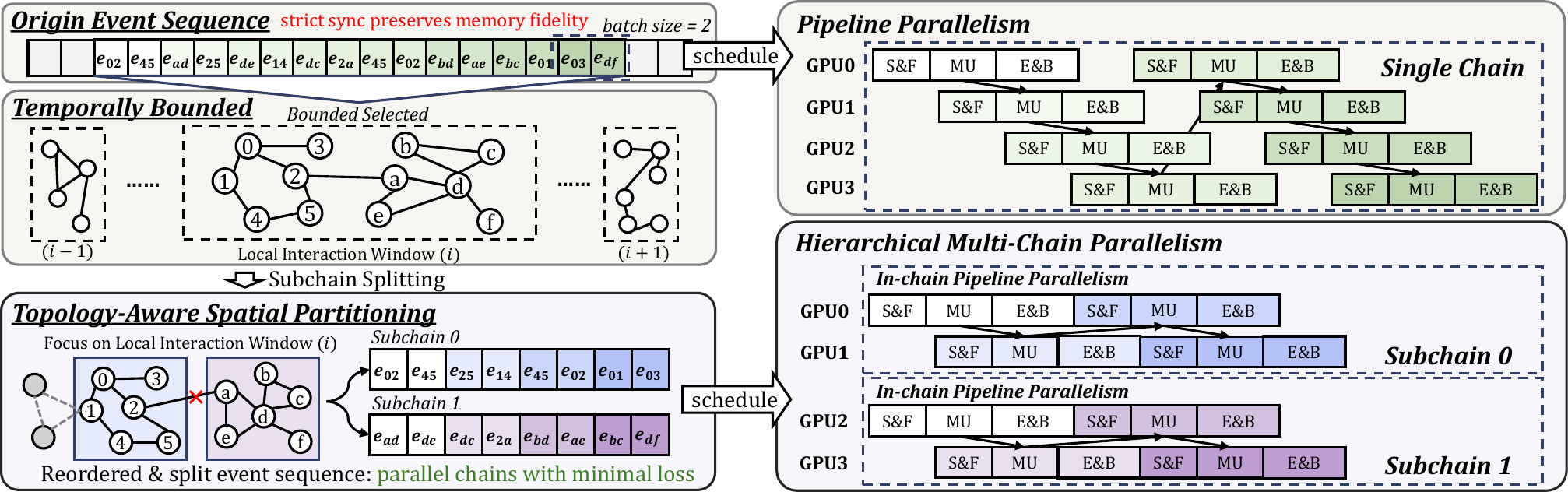}
\caption{Comparison between the original pipeline parallelism and Topology-Aware Hierarchical Multi-Chain Parallelism. Using eight consecutive batches (batch size=2) as an example, the original design executes batches in a single pipeline and incurs frequent unnecessary synchronization. Our design partitions events into multiple topology-aware sub-chains and maps each sub-chain to a GPU group for parallel execution across sub-chains with minimal dependency loss.
}
\label{fig:multi-chain}
\end{figure*}

\Solved{\Yue{
Maybe we should think about reordering the section into: partition -> batch -> execution? This sounds more organized from top to bottom.
}}

% \Yue{for the following: Clarity is the first priority. Note (i) avoid use before define in any term. (ii) from top to bottom, always clear intuition or goals first, then details, then the reviewer can start their imagination. 
% }
% \subsection{Topology-Aware Hierarchical Multi-Chain Parallelism}
\subsection{Temporally Bounded Spatial Partitioning}

Motivated by \textbf{Observation~1}, we propose \textit{Temporally Bounded Spatial Partitioning} to improve TGNN training efficiency by reducing unnecessary synchronization while preserving critical memory dependencies. The key idea is to organize the execution hierarchically. We first partition the input event stream along the temporal dimension to bound memory synchronization (i.e., enforce necessary memory synchronization between windows), then partition each temporal window along the spatial dimension into sub-chains, and finally execute the resulting sub-chains in parallel with synchronization only at window boundaries.

\textbf{Temporal partitioning for bounded memory synchronization.}
% \color{olive}
We first divide the input event stream $\mathcal{E}={e_i}_{i=1}^{T}$ into consecutive temporal windows $\mathcal{W}_1, \mathcal{W}_2,\ldots,\mathcal{W}_M$, where each $\mathcal{W}_m$ contains a contiguous segment of events.
% \color{black}
The purpose of this step is to bound how long memory dependencies can remain unsynchronized. As discussed in Section~\ref{subsec: motivation}, if spatial partitioning is applied over an excessively long event range, omitted cross-partition dependencies accumulate and significantly degrade training quality.
By restricting dependency relaxation within bounded temporal windows, \OurMethod{} preserves necessary synchronization across time while preventing long-range dependency loss from growing uncontrollably. 
In other words, temporal partitioning defines the scope within which dependencies may be relaxed, while still enforcing synchronization between adjacent windows.
We empirically set the window size as $80,000$ events by default, and quantitatively evaluate different window sizes in Section~\ref{subsec: eval_sens}.

% We first divide the input event stream into consecutive temporal windows and execute these windows in chronological (i.e., sequential) order. 
% The purpose of this step is to bound how long memory dependencies can remain unsynchronized. As discussed in Section~\ref{subsec: motivation}, if spatial partitioning is applied over an excessively long event range, omitted cross-partition dependencies accumulate and significantly degrade training quality.
% By restricting execution to bounded temporal windows, we preserve necessary memory synchronization across time while preventing long-range dependency loss from growing uncontrollably.
% In other words, temporal partitioning defines the scope within which dependencies may be relaxed, while still enforcing synchronization between adjacent windows.
% We empirically set the window size as $80,000$ events by default, and quantitatively evaluate different window sizes in Section~\ref{subsec: eval_sens}.

\textbf{Spatial partitioning for topology-aware parallelism.}
% \color{olive}
Within each temporal window $\mathcal{W}_m$, we build a local interaction graph $G_m=(V_m,E_m,w^V_m,w^E_m)$. A node $v\in V_m$ appears if it participates in at least one event in $\mathcal{W}_m$. An edge $(u,v)\in E_m$ exists if any event occurs between $u$ and $v$ during this window. We assign each edge a weight $w^E_m(u,v)$ equal to the number of events observed between the two nodes, and assign each node a weight $w^V_m(v)$ equal to the number of events involving that node. 
% We then partition $G_m$ into $P$ balanced subgraphs by solving:
% \begin{displaymath}
% \min*{\pi_m:V_m\rightarrow {1,\ldots,P}}
% \end{displaymath}
% \begin{displaymath}
% \sum*{(u,v)\in E_m} w^E_m(u,v)\cdot \mathbf{1}[\pi_m(u)\neq \pi_m(v)]
% \end{displaymath}
% subject to
% \[
% \sum_{v:\pi_m(v)=p} w^V_m(v)
% \le
% (1+\delta)\frac{\sum_{v\in V_m}w^V_m(v)}{P},
% \quad \forall p\in{1,\ldots,P}.
% \]
We then partition $G_m$ into $P$ subgraphs. Let $\pi_m(v)\in\{1,\ldots,P\}$ denote the subgraph assigned to node $v$. The goal is to minimize the weighted cut:
\[
\operatorname{Cut}(\pi_m)
=
\sum_{(u,v)\in E_m:\pi_m(u)\neq \pi_m(v)}
w^E_m(u,v),
\]
while keeping the total node weight of each subgraph approximately balanced:
\[
\sum_{v:\pi_m(v)=p} w^V_m(v)
\le
(1+\delta)\frac{\sum_{v\in V_m}w^V_m(v)}{P},
\quad \forall p\in\{1,\ldots,P\}.
\]
Here, $\delta$ controls the allowed load imbalance. The weighted cut captures the amount of interaction split across different subgraphs, which approximates cross-sub-chain memory dependencies. The balance constraint prevents skewed partitions from creating GPU workload imbalance.
% In our implementation, we use METIS~\cite{karypis1997metis} to solve this weighted graph partitioning problem efficiently. Since a larger edge weight indicates more frequent interaction, this partitioning tends to place strongly coupled nodes in the same subgraph. As a result, most memory dependencies are kept within individual subgraphs. We finally convert the events associated with each subgraph back into a chronological event sequence, which we call a \textit{subchain}. Events inside each subchain preserve their original temporal order, while different subchains are only weakly coupled.
In our implementation, we use METIS~\cite{karypis1997metis} to solve this weighted graph partitioning problem efficiently. Since a larger edge weight indicates more frequent interaction, this partitioning tends to place strongly coupled nodes in the same subgraph. As a result, most memory dependencies are kept within individual subgraphs. We finally convert the events associated with each subgraph back into a chronological event sequence, which we call a \textit{subchain}. Events inside each subchain preserve their original temporal order, while different subchains are only weakly coupled. 
% Although we use METIS in our implementation, the design is not tied to a specific partitioner; other topology aware heuristics, such as greedy assignment or streaming graph partitioning, can also be used as long as they reduce cross subchain interactions while maintaining load balance.
The proposed design is compatible with other lightweight topology-aware strategies, such as greedy assignment or streaming graph partitioning, which can incorporate additional partitioning factors, including node degrees, dependency strength, and event-load balance, to further reduce cross-subchain interactions.
% \color{black}
% Within each temporal window, we build a local interaction graph from the events in that window. 
% This graph contains all nodes that appear in the window, and connects two nodes if any events occur between them during the window. 
% For each edge, we assign a weight equal to the number of events observed between the two nodes; for each node, we assign a weight equal to the number of events involving that node. 
% We then use a topology-aware partitioner such as METIS~\cite{karypis1997metis} to divide this graph into multiple balanced subgraphs with similar total node weights, while minimizing the total weight of edges split across different subgraphs.
% Because a larger edge weight indicates more frequent interaction, this partitioning tends to keep strongly interacting nodes in the same subgraph. 
% As a result, most dependencies remain within individual subgraphs rather than across them. We finally convert each subgraph back into an event sequence, which we call a \textit{sub-chain}, so that different sub-chains are only weakly coupled and can be processed more independently.

\textbf{Partitioning-based bounded parallelism.}
After temporal and spatial partitioning, different sub-chains within the same temporal window are assigned to different GPU groups for concurrent execution, as illustrated in \cref{fig:multi-chain}.
Within each sub-chain, the original TGNN training order is preserved, so batches are still executed sequentially. 
Because spatial partitioning has already minimized cross-chain coupling, synchronization across sub-chains is eliminated.
The system synchronizes node memories only after all sub-chains in the current window complete, before moving to the next temporal window. 
As a result, \OurMethod{} replaces a single global execution pipeline with multiple local pipelines that progress in parallel, reducing communication overhead and shortening dependency-induced stalls.

% \Yue{we can remove this if space is short, now it feels a little bit far-fetched---they are actually orthogonal. Also, we already have the overview at the beginning that glues them together. (The paragraph has now been annotated)}

% Overall, \textit{Temporally Bounded Spatial Partitioning} improves scalability in two ways. First, it exposes more parallelism by allowing multiple weakly coupled sub-chains to execute simultaneously. Second, it reduces synchronization overhead by limiting memory synchronization to temporal window boundaries, rather than requiring fine-grained synchronization throughout the entire execution. This design is also complementary to \textit{Intra-Batch Memory Refinement}: while memory refinement recovers missing short-range dependencies within each execution unit, temporally bounded spatial partitioning reduces unnecessary synchronization across execution units.
% \color{black}

\begin{figure*}[!t]
\centering
\includegraphics[width=\linewidth]{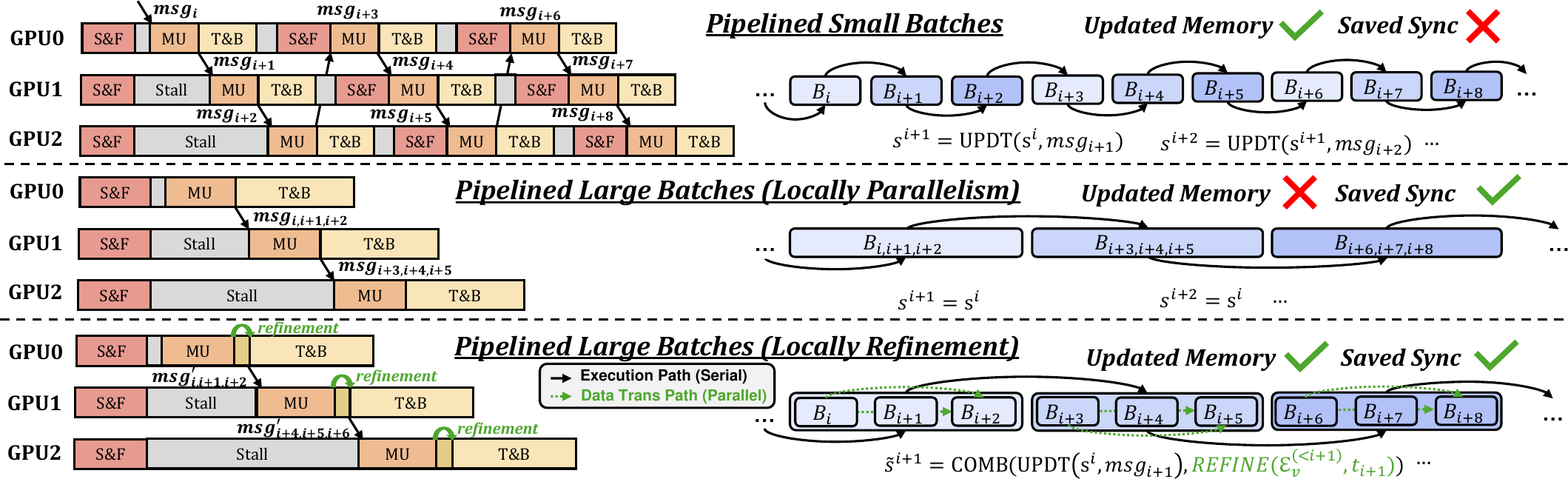}
\caption{
% Comparison of three multi-GPU pipeline execution strategies for M-TGNN training. \textit{(a) Pipelined Small Batches}: each GPU processes one small batch per pipeline stage, preserving full memory continuity at the cost of frequent cross-GPU synchronization. \textit{(b) Pipelined Large Batches (Locally Parallelism)}: multiple small batches are merged and their micro-batches processed in parallel, reducing synchronization but using stale memory from the previous large batch for all micro-batches. \textit{(c) Pipelined Large Batches (Locally Refinement)}: same large-batch structure as (b), but a lightweight memory refinement step (green) propagates short-range dependencies among micro-batches, recovering memory quality without cross-GPU synchronization. \includegraphics[width=.7em]{picture/label/label1.pdf}: sample \& fetch; \includegraphics[width=.7em]{picture/label/label2.pdf}: memory update; \includegraphics[width=.7em]{picture/label/label3.pdf}: memory refinement; \includegraphics[width=.7em]{picture/label/label4.pdf}: training \& backward.
Comparison of three multi-GPU pipeline execution strategies for TGNN training. \textit{(a) Pipelined Small Batches}: each GPU executes one small batch per stage in strict sequential order. \textit{(b) Pipelined Large Batches (Local Parallelism)}: micro-batches within each large batch are scheduled in parallel, each consuming the same stale memory from the previous batch. \textit{(c) Pipelined Large Batches (Local Refinement)}: same scheduling as (b), but a refinement step (green) is inserted between micro-batches to propagate short-range dependencies locally before the next micro-batch executes.
% Black arrows denote serial execution paths; green dashed arrows denote data usage paths that proceed in parallel.
\includegraphics[width=.7em]{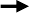}: actual scheduling dependencies that enforce \textbf{serial execution} order;
\includegraphics[width=.7em]{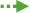}: data usage relationships among micro-batches, which allow \textbf{parallel execution};
\includegraphics[width=.7em]{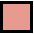}: sample \& fetch;
\includegraphics[width=.7em]{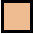}: memory update;
\includegraphics[width=.7em]{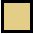}: memory refinement;
\includegraphics[width=.7em]{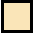}: training \& backward.
\Solved{\Yang{Yes, this understanding is correct. Now, the third sub-figure maybe need to be edited, because it seems that $B_i, B_{i+1}, B_{i+2}$ will be scheduled to the same GPU and run "memory update, memory refine" simultaneously. Here, as we last discussed, we have no need to place the detail of our refinement strategy here, and move it to the "detail" section.} \Yue{Overall, this is clear now. But two more things may help better: (1) for refinement: the red arrows seems serial, think about making it parallel like previous version could help distinguish---no arrows like $B_i\rightarrow B_{i+1}\rightarrow B_{i+2}$. (2) The equations are confusing for now....are you what to describe what happened in one single batch?}\Yang{(1) Maybe previous version can't describe the refinement in a batch. We can talk about how to show it. (2) Yes, later to talk.}}
}
\label{compare}
\end{figure*}

\subsection{Intra-batch Memory Refinement}\label{subsec: design_mem_refine}

\yuepre{
Motivated by \textbf{Observation~1}, we propose \emph{Intra-Batch Memory Refinement}, as illustrated in \cref{compare}(c), to achieve the efficiency of large-batch execution without losing the fresh memory states provided by small batches. 
As shown in \cref{compare}(a), small-batch execution preserves frequent memory updates between neighboring batches, but suffers from frequent cross-device synchronization and inter-batch pipeline stalls. In contrast, straightforward large-batch execution in \cref{compare}(b) reduces synchronization frequency but delays memory updates too much and therefore hurts model accuracy. Our goal is to combine the advantages of both: process a large batch efficiently while keeping node memories as up-to-date as if the batch had been executed as several small batches with intermediate memory updates in between. 
To this end, we propose \emph{Intra-Batch Memory Refinement} consists of three steps: mini-batch split, parallel memory refinement, and node-memory aggregation.
}

% \yueafter{
Motivated by \textbf{Observation~2}, we propose \textit{Intra-Batch Memory Refinement}, as illustrated in \cref{compare}(c). Compared with small-batch execution in \cref{compare}(a), which preserves frequent memory updates but incurs frequent synchronization and pipeline stalls, and large-batch execution in \cref{compare}(b), which improves efficiency but uses overly stale node memories, our design combines the strengths of both. It retains the efficiency of large-batch execution while keeping node memories nearly as fresh as if the batch had been processed as several small batches with intermediate memory updates.
% }

\textbf{Micro-batch Split.}
Instead of treating a large batch as an indivisible unit, we divide it into an ordered sequence of micro-batches,
$B_t = \{\mathcal{M}_1, \mathcal{M}_2, \ldots, \mathcal{M}_K\}$,
where each micro-batch $\mathcal{M}_k$ contains the same number of events as the original small-batch baseline, and the micro-batches remain ordered by event timestamp. In this way, the large batch retains the temporal structure of several small batches, while still preserving the lower synchronization frequency of large-batch execution.

\textbf{Parallel Memory Refinement.}
To avoid sequential updates between micro-batches, which cause frequent stalls and synchronizations, we allow micro-batches to execute in parallel and use a lightweight refinement module to approximate the effect of the omitted intermediate updates.
Specifically, when processing micro-batch $\mathcal{M}_i$, we first apply the standard memory update using the available but stale node memories at the beginning of the large batch, and then refine that memory using the most recent events from preceding micro-batches $\mathcal{M}_1,\ldots,\mathcal{M}_{i-1}$. This refinement compensates for the skipped intermediate updates without requiring cross-device synchronization.
Formally, let $\mathbf{s}_v^{-}$ denote the committed memory of node $v$ before batch $B_t$, let $\mathbf{m}_v^{(i)}$ denote the message aggregated for $v$ in micro-batch $\mathcal{M}_i$, and let $\mathcal{E}_v^{(<i)}$ denote the most recent events involving $v$ from preceding micro-batches. The refined memory of $v$ in $\mathcal{M}_i$ is computed as
\begin{equation}
\mathbf{s}_v^{(i)}
=
\mathrm{COMB}\!\left(
\mathrm{UPDT}\!\left(\mathbf{s}_v^{-}, \mathbf{m}_v^{(i)}\right),
\mathrm{REFINE}\!\left(\mathcal{E}_v^{(<i)}, t_i\right)
\right),
\label{eq:memory_refine}
\end{equation}
% \zypre{
% where $\mathrm{UPDT}(\cdot)$ is the standard memory update operator and $\mathrm{REFINE}(\cdot)$ is a lightweight refinement module, such as an RNN or a linear projector. The first term captures long-range historical context from the previous large batch, while the second term restores short-range temporal influence that would otherwise be missed due to skipped intermediate updates.
% }
% \zyafter{
where $\mathrm{UPDT}(\cdot)$ is the standard memory update operator and $\mathrm{REFINE}(\cdot)$ is a lightweight refinement module, such as an RNN or a linear projector. The first term captures long-range context from the previous large batch, while the second restores short-range temporal influence missed by skipped intermediate updates.
% }

\textbf{Node-memory aggregation.}
After all micro-batches in $B_t$ have been processed, we aggregate the refined node memories to produce the final memory state that will be consumed by subsequent batches. This step is necessary because the same node may appear in multiple micro-batches, leading to multiple refined memory replicas within the same large batch. 
% Following common aggregation operators used in prior TGNN designs~\cite{kumar2019predicting,rossi2020temporal}, we consider two strategies: \emph{latest memory}, which keeps the replica with the most recent timestamp, and \emph{average memory}, which averages all replicas produced within the batch.
Following common aggregation operators used in prior TGNN designs~\cite{kumar2019predicting,rossi2020temporal}, we adopt the \emph{latest memory} strategy, which keeps the replica with the most recent timestamp among all replicas produced within the batch.
% \textcolor{red}{We evaluate different options in Section XXX.}
% \color{black}
% \Yue{do we have sensitivity study on different aggregate schemes? if not just one.} \Yang{Yes, I will do exps on different aggregate schemes.}

\subsection{Critical-Path Optimization for Multi-GPU Pipeline Execution}

\Solved{\Yue{No need to introduce equations and symbols if they do not help to explain figures/algorithms---in ML paper we need this but here not.} \Yang{Get!}}

As discussed in \textbf{Observation~3}, a downstream GPU often cannot proceed until the required memory updates from the previous GPU have been both produced and transferred. As the pipeline deepens, these dependency-critical operations become harder to hide and increasingly create pipeline stalls. 
To address this problem, we optimize multi-GPU pipeline execution from two aspects. 
First, we remove non-urgent memory updates from the critical path via \textit{Priority-aware memory-update decoupling}.
Second, we move repetitive preprocessing work out of latency-sensitive stages and overlap it with other operations via \textit{Critical-path precomputation and overlap}. Together, these optimizations shorten the effective critical path and reduce pipeline stalls.

\begin{figure}[thbp]
\centering
\includegraphics[width=\linewidth]{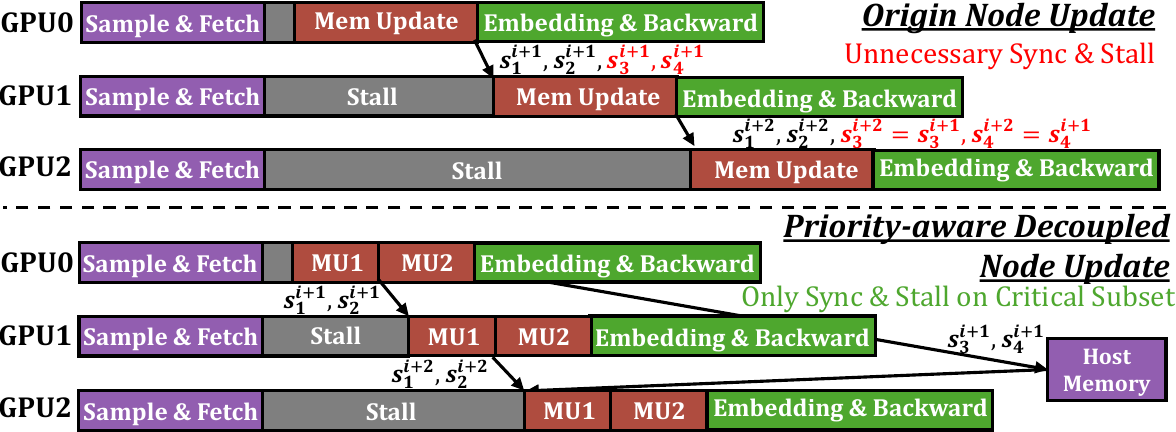}
\caption{Comparisons between unified node update and priority-aware decoupled node update.}
\label{fig:decouple}
\end{figure}

\textbf{Priority-aware memory-update decoupling.}
Recalling \textbf{Observation~3}, only a small subset of memory updates is immediately required by the next dependent batch, while many others are consumed later or not revisited in the near term. 
To avoid unnecessary waiting cost, we split the updates produced by each batch into two groups: \textit{urgent} updates and \textit{non-urgent} updates. 
Then we handle their execution differently, as illustrated in \cref{fig:decouple}:
Urgent updates are those immediately needed by the next dependent batch, so they are updated and transferred first via GPU--GPU communication (e.g., NVLink). This allows the downstream GPU to resume execution as soon as the required memory state becomes available. 
In contrast, non-urgent updates are removed from the critical path. They are handled separately without blocking the execution of the next batch on remote GPUs, and the resulting memory will first be written back to the CPU and then loaded onto remote GPUs on demand. 
This design preserves correctness for dependency-critical updates while significantly reducing the amount of communication that lies on the critical path. 
% As a result, it shortens synchronization stalls, improves overlap between computation and communication, and reduces pipeline bubbles in multi-GPU execution.

\textbf{Critical-path precomputation and overlap.}
Beyond memory transfer itself, multi-GPU TGNN execution also incurs repeated overhead in preparing memory accesses. 
Specifically, before a batch can fetch the required node memories, the system often performs preprocessing steps such as node-index deduplication, sorting, remapping, request construction, and buffer packing~\cite{zhou2023disttgl}, because the required memories may reside in different locations, such as CPU memory or GPU memory on different devices. 
Although each step is lightweight in isolation, repeatedly performing them on the critical path adds noticeable delay and further prolongs pipeline stalls.

However, much of this preprocessing depends only on information that is already known before execution begins. In particular, once the event sequence, neighbor sampling results, and partition assignments are determined, the set of accessed nodes and the corresponding memory-access patterns are also fixed. 
This means that many metadata structures, such as accessed-node lists, index remapping tables, communication offsets, and data movement plans, can be prepared in advance rather than rebuilt repeatedly at runtime and stalled by node memory synchronizations.

Therefore, we move this preprocessing work off the critical path by precomputing it before receiving updated node memories from the previous batch, hiding its overhead behind the memory-update critical path.
Specifically, while a GPU waits for updated memory from the previous batch or fetches the node features needed by the current batch, \OurMethod{} concurrently prepares the metadata needed for future batches. In our implementation, we also explore Python 3.13's free-threading mode to implement such thread-level parallelism more easily, allowing I/O data preparation to run concurrently with latency-sensitive training stages.

% Overall, this optimization shortens the effective critical path and improves GPU utilization by increasing overlap among preprocessing, communication, and computation.
% \color{black}
\section{Evaluation}
\begin{figure*}[t]
\centering
\includegraphics[width=0.95\linewidth]{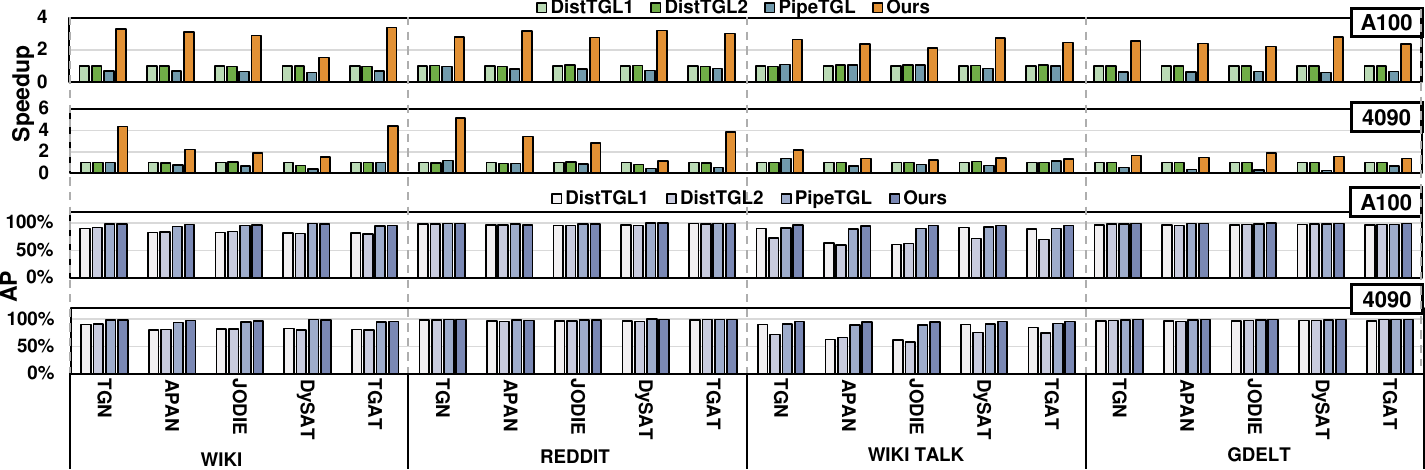}
\caption{
% Overall performance comparison of \OurMethod{} against \textsc{DistTGL1}, \textsc{DistTGL2}, and \textsc{PipeTGL} on two single-node 8-GPU platforms, NVIDIA A100 and RTX 4090, across four datasets and three TGNN models. (a) End-to-end training speedup relative to \textsc{DistTGL1}. (b) Average precision (AP) of the trained models.
Overall performance comparison on NVIDIA A100 and RTX 4090 across four datasets and five TGNN models. The first two rows show end-to-end speedup over \textsc{DistTGL}$_1$, and the last two rows show average precision (AP).
}
\label{fig:exp-overall}
\end{figure*}
\subsection{Methodology}

% \textbf{Models.} We evaluate \OurMethod{} using three representative TGNN models as follows: (a) JODIE~\cite{kumar2019predicting}, which models dynamic interactions with coupled recurrent updates and time-aware projection.
%  (b) TGN~\cite{rossi2020temporal}, which maintains per-node memories and combines recurrent memory updates with temporal graph embedding.
%  (c) APAN~\cite{wang2021apan}, which performs asynchronous mail propagation and attention-based temporal embedding.
% We follow the model configuration used in \textsc{PipeTGL}~\cite{liu2025pipetgl} as shown in \cref{tab:model-config}.

\textbf{Models.}
% \color{olive}
We evaluate \OurMethod{} on five representative TGNN models, including three CTDG-based models: (1) JODIE~\cite{kumar2019predicting} uses recurrent updates with time-aware projection, (2) TGN~\cite{rossi2020temporal} combines node memory with temporal graph embedding, and (3) APAN~\cite{wang2021apan} adopts asynchronous mail propagation and attention-based embedding; 
and DTDG-based models to assess adaptability: (1) DySAT~\cite{sankar2020dysat}, which models temporal evolution across graph snapshots with RNNs, and (2) TGAT~\cite{xu2020inductive}, which combines temporal encoding with attention-based neighborhood aggregation.
We follow \textsc{Cascade}~\cite{dai2025cascade} configurations, as shown in \cref{tab:model-config}.
% \color{black}

\begin{table}[!ht]
    \centering
    \caption{Details of TGNN models, with $d_{out}=100$}
    \footnotesize
    \begin{tabular}{cccc}
    \hline
        \textbf{} & \textbf{Sample Message} & \textbf{Memory Update} & \textbf{Node Embedding} \\ \hline
        \textbf{JODIE} & most recent (n=1) & RNN & Identity \\ 
        \textbf{TGN} & most recent (n=1) & GRU & GAT \\ 
        \textbf{APAN} & most recent (n=10) & Transformer & Identity \\
        \hline
        \textbf{TGAT} & uniform (n=10) & Identity & 2-layer GAT \\ 
        \textbf{DySAT} & uniform (n=10) & GAT & RNN \\
        \hline
    \end{tabular}
    \label{tab:model-config}
\end{table}

\textbf{Datasets.}
% We evaluate \OurMethod{} on four real-world temporal graph datasets with diverse scales and characteristics, including Wikipedia (\texttt{WIKI}), Reddit (\texttt{REDDIT})~\cite{kumar2019predicting}, Wikipedia Talk (\texttt{WIKI-TALK})~\cite{leskovec2010signed}, and GDELT (\texttt{GDELT})~\cite{zhou2022tgl}. 
% \texttt{WIKI} and \texttt{REDDIT} are relatively small-scale datasets that are widely used in prior TGNN studies. 
% \texttt{WIKI-TALK} is a large-scale temporal interaction graph with millions of nodes and edges, which allows us to assess the effectiveness of \OurMethod{} on substantially larger workloads. 
% \texttt{GDELT} is an event graph constructed from news records and contains nearly 0.2 billion temporal interactions, serving as a stress test for scalability under event-intensive settings. 
% For datasets without original edge or node features, we generate synthetic features following prior practice~\cite{dai2025cascade,zhou2022tgl}. The detailed dataset statistics are summarized in Table~\ref{tab:dataset-stat}. Random features are denoted by ``*'' in the table.
We evaluate \OurMethod{} on four real-world temporal graph datasets of diverse scales: Wikipedia (\texttt{WIKI}), Reddit (\texttt{REDDIT})~\cite{kumar2019predicting}, Wikipedia Talk (\texttt{WIKI-TALK})~\cite{leskovec2010signed}, and GDELT (\texttt{GDELT})~\cite{zhou2022tgl}. 
\texttt{WIKI} and \texttt{REDDIT} are small-scale benchmarks widely used in prior TGNN studies, \texttt{WIKI-TALK} is a large-scale interaction graph with millions of nodes and edges, and \texttt{GDELT} contains nearly 0.2 billion temporal interactions, serving as a stress test for scalability.
For datasets without original node or edge features, we generate synthetic features following prior practice~\cite{dai2025cascade,zhou2022tgl}. Detailed statistics are in Table~\ref{tab:dataset-stat}, where random features are marked by ``*''.

\begin{table}[!ht]
    \centering
    \small
    \caption{Dataset statistics. $\lvert d_v\rvert$ and $\lvert d_e\rvert$ show the dimensions of node features and edge features, respectively.}
    \begin{tabular}{ccccc}
    \hline
        \textbf{Dataset} & \textbf{$\lvert V\rvert$} & \textbf{$\lvert E\rvert$} & \textbf{$\lvert d_v\rvert$} & \textbf{$\lvert d_e\rvert$} \\ \hline
        \textbf{WIKI} & 9,227 & 157,474 & 172 & 100* \\ 
        \textbf{REDDIT} & 11,000 & 672,447 & 172 & 100* \\ 
        \textbf{WIKI-TALK} & 1,140,149 & 7,833,140 & 32* & 100* \\ 
        \textbf{GDELT} & 16,682 & 191,290,882 & 186 & 413 \\ \hline
    \end{tabular}
    \label{tab:dataset-stat}
\end{table}

\textbf{Environment.}
% We conduct all experiments on a single machine with eight NVIDIA GeForce RTX 4090 GPUs, an Intel Xeon Platinum 8375C CPU, and 512\,GB of system memory. The system runs Ubuntu 22.04 with CUDA 12.6\cite{nickolls2008scalable}, Python 3.13 (free-threaded), PyTorch 2.9.0\cite{wang2019deep}, and DGL\cite{paszke2019pytorch} 2.5. For multi-GPU execution, we use the distributed package in PyTorch with NCCL as the communication backend. All methods are evaluated under the same hardware and software environment to ensure fair comparison.
We conduct all experiments on two single-node 8-GPU platforms: one equipped with eight NVIDIA GeForce RTX 4090 GPUs, and the other with eight NVIDIA A100 GPUs. Both platforms use an Intel Xeon Platinum 8375C CPU and 512\,GB of system memory. The system runs Ubuntu 22.04 with CUDA 12.6\cite{nickolls2008scalable}, Python 3.13 (free-threaded), PyTorch 2.9.0\cite{paszke2019pytorch}, and DGL 2.5\cite{wang2019deep}. For multi-GPU execution, we use the distributed package in PyTorch with NCCL as the communication backend. All methods are evaluated under the same hardware and software environment to ensure fair comparison.

\textbf{Compared Methods.}
Our experiments compare the following representative multi-GPU TGNN training systems:
\squishlist
\item \textsc{PipeTGL}~\cite{liu2025pipetgl} (baseline) is a pipeline-parallel system that preserves memory continuity across batches and reduces pipeline bubbles through scheduling and communication optimizations. 
\item \textsc{DistTGL}~\cite{zhou2023disttgl} is a distributed training framework that improves scalability via tailored parallel training strategies and system optimizations. 
In multi-GPU scenarios, we evaluate \textsc{DistTGL} with batch parallelism (denoted as \textsc{DistTGL}$_1$) and epoch parallelism (denoted as \textsc{DistTGL}$_2$).
\squishend

\subsection{Overall Performance}

% \color{olive}
We first evaluate the end-to-end performance of our method against three representative baselines, on two hardware platforms (A100 and RTX 4090), four datasets, and five representative TGNN models. As shown in \cref{fig:exp-overall}, our method consistently delivers the best overall efficiency--accuracy trade-off across all evaluated settings.

\textbf{Training speedup.}  
% As shown in \cref{fig:exp-overall}(a), our method consistently outperforms all baselines in throughput. Across all 24 evaluated configurations, it achieves the highest speedup, ranging from \textbf{1.25$\times$} to \textbf{5.16$\times$}, with an average of \textbf{2.63$\times$}.
% On sparse temporal graphs such as \texttt{WIKI} and \texttt{WIKI-TALK}, \OurMethod{} outperforms all baselines in both throughput and accuracy. On \texttt{WIKI}, it achieves an average speedup of \textbf{2.97$\times$} over \textsc{DistTGL} and \textbf{3.95$\times$} over \textsc{PipeTGL} across all models and platforms. On \texttt{WIKI-TALK}, it achieves \textbf{1.99$\times$} over \textsc{DistTGL} and \textbf{1.95$\times$} over \textsc{PipeTGL}. On denser and larger-scale datasets such as \texttt{REDDIT} and \texttt{GDELT}, \OurMethod{} delivers even higher speedups over \textsc{PipeTGL}, averaging \textbf{3.76$\times$} on \texttt{REDDIT} and \textbf{4.03$\times$} on \texttt{GDELT}, with a peak of \textbf{5.16$\times$}.
As shown in \cref{fig:exp-overall}, \OurMethod{} consistently outperforms all baselines in throughput.
Across all 40 evaluated configurations, it achieves the highest speedup, ranging from \textbf{1.15$\times$} to \textbf{5.16$\times$}, with an average of \textbf{2.84$\times$}.
On sparse temporal graphs such as \texttt{WIKI} and \texttt{WIKI-TALK}, \OurMethod{} achieves average speedups of \textbf{2.90$\times$} and \textbf{1.96$\times$} over \textsc{DistTGL}, and \textbf{3.95$\times$} and \textbf{2.05$\times$} over \textsc{PipeTGL}, respectively.
On denser and larger-scale datasets such as \texttt{REDDIT} and \texttt{GDELT}, \OurMethod{} achieves \textbf{3.17$\times$} and \textbf{2.03$\times$} speedups over \textsc{DistTGL}, while outperforming \textsc{PipeTGL} by \textbf{3.91$\times$} and \textbf{4.02$\times$}, respectively.
The consistent gains across both CTDG and DTDG models demonstrate the generality of \OurMethod{} across different temporal modeling paradigms.

% \textbf{Model quality.} 
% \OurMethod{} reaches the highest average precision (AP), \textbf{95.1\%}, among all evaluated methods.  On sparse temporal graphs, AP is more sensitive to memory discontinuity because each event provides less redundant context and skipped dependencies are harder to recover. 
% On \texttt{WIKI-TALK}, \OurMethod{} achieves \textbf{95.1\%} AP, surpassing \textsc{PipeTGL} and \textsc{DistTGL} by \textbf{5.1} and \textbf{23.7} percentage points, respectively. On \texttt{WIKI}, \OurMethod{} achieves \textbf{97.3\%} AP, surpassing \textsc{PipeTGL} by \textbf{1.6} and \textsc{DistTGL} by \textbf{12.1} percentage points. On denser datasets, \OurMethod{} reaches \textbf{99.5\%} AP on \texttt{GDELT}, surpassing \textsc{PipeTGL}'s \textbf{98.6\%}, and remains competitive with \textsc{PipeTGL} on \texttt{REDDIT} (\textbf{98.1\%} vs.\ \textbf{98.2\%}). 
% The accuracy gain mainly comes from DepBloom's dependency-aware execution. 
% Temporally bounded spatial partitioning keeps strongly coupled events within the same sub-chain. Intra-batch memory refinement recovers recent temporal dependencies that are lost in enlarged batches. Together, these mechanisms reduce synchronization frequency while maintaining the memory freshness required for accurate prediction.

\textbf{Model quality.}
\OurMethod{} achieves the highest average precision (AP) of \textbf{97.6\%} across all evaluated configurations, compared with \textbf{96.0\%} for \textsc{PipeTGL}, and obtains the highest AP in \textbf{35 of 40} configurations.
The advantage is particularly clear on \texttt{WIKI-TALK}, where \OurMethod{} achieves an average AP of \textbf{95.3\%}, compared with \textbf{90.7\%} for \textsc{PipeTGL}.
It also improves average AP from \textbf{96.1\%} to \textbf{97.0\%} on \texttt{WIKI} and from \textbf{98.4\%} to \textbf{99.4\%} on \texttt{GDELT}, while remaining comparable on \texttt{REDDIT} (\textbf{98.7\%} vs.\ \textbf{98.6\%}).
The quality gains mainly come from DepBloom's dependency-aware execution: temporally bounded spatial partitioning keeps strongly coupled events within the same sub-chain, while intra-batch memory refinement recovers recent dependencies lost in enlarged batches.
Together, these mechanisms expose more parallelism while preserving the memory freshness required for prediction.

\textbf{Cross-platform trends.}
DepBloom delivers consistent gains across both hardware platforms.
On A100 and RTX 4090, it achieves average speedups of
\textbf{2.97$\times$} and \textbf{2.70$\times$} across all baseline comparisons, respectively. The slightly lower gain on RTX 4090 mainly results from its more limited memory capacity, which constrains achievable batch sizes on large-scale datasets. AP improvements remain consistent across platforms, showing that the benefits of our design are not tied to a specific GPU architecture.
% \textbf{Model quality.}
% % On sparse temporal graphs, 例如 WIKI and WIKI_TALK, accuracy is more sensitive to memory discontinuity, because each event carries less redundant local context and the latest node state is harder to recover once a temporal dependency is skipped. 因此 disttgl 中的方法无法取得更高的收敛进度，平均 xx；而在 pipetgl 中，捕捉了更多依赖获取了更高的精度，却受到了频繁的 memory sync 开销难以有很好的性能。我们的方法在取得平均 xx 精度的同时，仍然保持了最高的性能， xxx。而在其他的数据集上，xxxx。

% \textbf{Cross-platform trends.} % 我们的方法在两种硬件平台上均能取得较高的性能。在 A100 上，由于具有更大的显存，我们可以更大地提高 batch size 以提高系统吞吐率，实现了平均 xxx 的加速。而在 RTX 4090 上，由于显存容量的限制，在较大的数据集上无法达到更大的 batch size，但仍能达到 xxx 加速，并取得更高的精度提升。这种一致性表明，我们的设计带来的收益并不依赖于特定的 GPU 架构，xxxxxxx。

% Overall, these results demonstrate that our method substantially improves multi-GPU TGNN training efficiency while maintaining, and often improving, model quality. This confirms that carefully relaxing non-critical dependencies and optimizing the execution path is more effective than either coarse-grained parallelization or pipeline-only optimization.

\begin{figure}[thbp]
\centering
\includegraphics[width=0.98\linewidth]{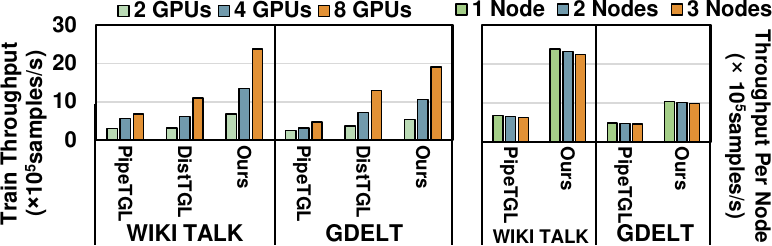}
\caption{Scalability on \texttt{WIKI-TALK} and \texttt{GDELT} with \texttt{APAN}, under single- and multi-node GPU configurations.
For multi-node, inter-node memory parallelism is combined with DepBloom's intra-node dependency-aware execution.
}
\label{fig:exp-scale}
\end{figure}

\subsection{Scalability}

% \color{olive}
We evaluate the scalability of \OurMethod{} with \texttt{APAN} on \texttt{WIKI TALK} and \texttt{GDELT}. For multi-node scaling, we combine inter-node memory parallelism with DepBloom's intra-node dependency aware execution on up to three RTX 4090 nodes. 
As shown in \cref{fig:exp-scale}, \OurMethod{} achieves \textbf{87.6\%}/\textbf{88.5\%} parallel efficiency at 8 GPUs on \texttt{WIKI-TALK}/\texttt{GDELT}. When scaling from $1\times8$ to $3\times8$ GPUs, it retains \textbf{94.3\%}/\textbf{95.1\%} of the per-node throughput, corresponding to \textbf{2.83$\times$}/\textbf{2.85$\times$} total-throughput scaling.

% The results show that \OurMethod{} preserves its intra-node efficiency when scaling across nodes. At $3\times8$ GPUs, it achieves \textbf{3.61$\times$}/\textbf{2.18$\times$} higher throughput than \textsc{PipeTGL} on \texttt{WIKI-TALK}/\texttt{GDELT}. 
% These results show that DepBloom preserves its intra-node efficiency when combined with inter-node memory parallelism.
% Consequently, increasing the number of nodes introduces only modest per-node throughput degradation while preserving the benefits of dependency-aware execution.
The results show that \OurMethod{} largely preserves its intra-node efficiency when combined with inter-node memory parallelism. Within each node, \OurMethod{} exposes parallel sub-chains while keeping most temporal dependencies localized, increasing useful execution concurrency without proportionally increasing cross-GPU synchronization. In contrast, \textsc{PipeTGL} remains constrained by cross-GPU memory synchronization as the pipeline scales, while \textsc{DistTGL} improves parallelism by relaxing the original dependency order, which can degrade model quality. Consequently, scaling to more nodes introduces only modest per-node throughput degradation while preserving the benefits of dependency-aware execution.
% \color{black}

\subsection{Optimization Analysis}

To understand the effectiveness of each optimization in \OurMethod{}, we perform an ablation study to quantify their individual contributions and examine how they improve end-to-end training performance.

% \textbf{Temporally Bounded Spatial Partitioning}

% We compare three settings in \cref{fig:exp-ab-split}: \textsc{W/O}, Time Bounded(\textsc{TB}), and Time Bounded+Spatial Split(\textsc{TB+SS}), where \textsc{TB} applies only temporal windowing and \textsc{TB+SS} further introduces topology-aware spatial partitioning inside each window. \textsc{TB} already improves the average AP from \textbf{91.24\%} to \textbf{92.10\%} and achieves \textbf{1.19$\times$} speedup over \textsc{W/O}. \textsc{TB+SS} further increases the average AP to \textbf{93.02\%} while maintaining a similar speedup of \textbf{1.19$\times$}. This shows that temporal bounding mainly contributes to efficiency, while spatial partitioning further improves model quality by preserving more useful local dependencies.

% \paragraph{Temporally Bounded Spatial Partitioning}
% We compare \textsc{W/O} and \textsc{W Temporally Bounded Spatial Partitioning} in \cref{fig:exp-ab-split}. Enabling temporally bounded spatial partitioning improves the average AP from \textbf{91.24\%} to \textbf{93.02\%} and delivers a \textbf{1.19$\times$} speedup compared with the baseline. 
% This result suggests that limiting partitioning to local temporal windows preserves useful short-range dependencies, while the spatial partitioning inside each window exposes additional parallelism.

\begin{figure}[thbp]
\centering
\includegraphics[width=\linewidth]{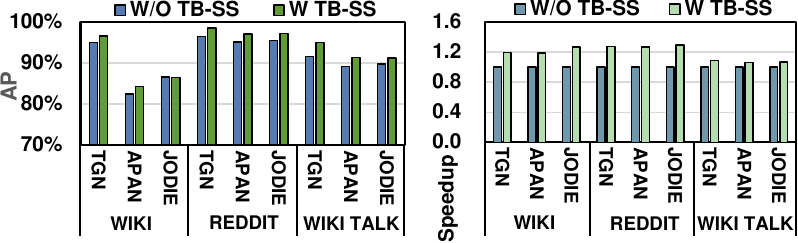}
\caption{Effect of temporally bounded spatial partitioning. (a) Average precision (AP). (b) End-to-end speedup.}
\label{fig:exp-ab-split}
\end{figure}

\textbf{Temporally Bounded Spatial Partitioning.}
We compare with (\textsc{W}) and without (\textsc{W/O}) Temporally Bounded Spatial Partitioning in \cref{fig:exp-ab-split}. With our optimization enabled, the average AP increases from \textbf{91.24\%} to \textbf{93.02\%}, together with a \textbf{1.19$\times$} speedup over the baseline. 
% \color{olive}
This result shows that jointly exploiting temporal and topological locality better balances the accuracy-efficiency tradeoff. Temporal bounds limit the accumulation of relaxed-dependency errors, while topology-aware partitioning preserves strongly coupled events within sub-chains, exposing parallelism across sub-chains and reducing cross-chain synchronization.
% \color{black}

% \textbf{Impact of intra-batch memory refinement.}
% We study \emph{intra-batch memory refinement} (IMR) under three settings: small batch(\textsc{SB}) with batch size 600, large batch(\textsc{LB}) with batch size 2400, and \textsc{LB+IMR(4)}, which splits each large batch into four refinement chunks of size 600. As shown in \cref{fig:exp-ab-refinement}, directly increasing the batch size consistently worsens validation loss, while IMR recovers most of the lost accuracy with only minor throughput overhead. Averaged in all six settings, the normalized loss increases to \textbf{1.32} under \textsc{LB} compared to \textsc{SB}, but drops to \textbf{1.03} with \textsc{LB+IMR(4)}; meanwhile, the average speedup changes only from \textbf{1.24$\times$} to \textbf{1.21$\times$}. The improvement is especially clear on \texttt{WIKI} and \texttt{REDDIT}, where \textsc{LB+IMR(4)} substantially reduces the loss increase introduced by \textsc{LB}. These results show that IMR effectively recovers short-range temporal dependencies ignored by large-batch training while preserving nearly all of its efficiency gain.

\begin{figure}[thbp]
\centering
\includegraphics[width=\linewidth]{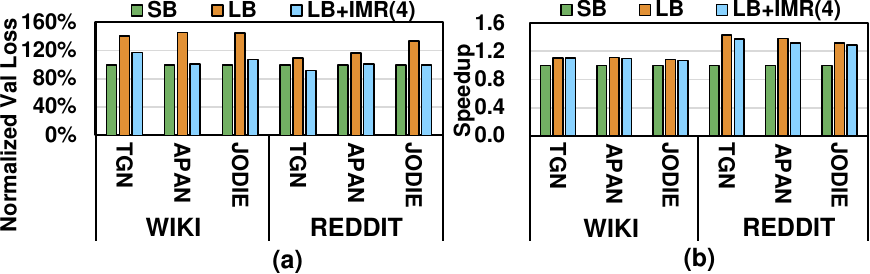}
\caption{Effect of intra-batch memory refinement. \textsc{SB} uses batch size 600, \textsc{LB} uses batch size 2400, and \textsc{LB+IMR(4)} splits each large batch into four chunks of size 600. (a) Normalized validation loss. (b) End-to-end speedup.}
\label{fig:exp-ab-refinement}
\end{figure}

\textbf{Impact of intra-batch memory refinement.}
We study \emph{intra-batch memory refinement} (IMR) under three settings: small batch (\textsc{SB}) with batch size 600, large batch (\textsc{LB}) with batch size 2400, and \textsc{LB+IMR(4)}, which splits each large batch into four chunks of size 600. As shown in \cref{fig:exp-ab-refinement}, \textsc{LB} consistently worsens validation loss, while IMR recovers most of the lost accuracy with little overhead. Averaged over all six settings, the normalized loss increases from \textbf{1.00} under \textsc{SB} to \textbf{1.32} under \textsc{LB}, but drops to \textbf{1.03} with \textsc{LB+IMR(4)}; meanwhile, the average speedup changes only from \textbf{1.24$\times$} to \textbf{1.21$\times$}. 
% These results show that IMR effectively restores short-range temporal dependencies ignored by large-batch training while preserving most of its efficiency gain.
% \color{olive}
These results show that large-batch training mainly loses accuracy due to the omission of short-range dependencies within the batch. IMR recovers the missing information while still avoiding costly sequential updates.
% \color{black}

% \paragraph{Critical-path optimization for multi-GPU pipeline execution.}
% We evaluate our critical-path optimization under the same \textbf{8$\times$A100} setup and batch-size configuration as the main experiment. As shown in \cref{fig:exp-ab-critical}, we break down the epoch time into memory synchronization (\textit{Memory Sync}), memory update on critical-path (\textit{Memory Update}), and the remaining execution time (\textit{Other Time}).
% This optimization consistently improves pipeline efficiency across all settings. Averaged over the three models, it achieves \textbf{1.11$\times$}, \textbf{1.22$\times$}, and \textbf{1.34$\times$} speedup on \texttt{WIKI}, \texttt{REDDIT}, and \texttt{WIKI-TALK}, respectively. The gain becomes larger on \texttt{REDDIT} and \texttt{WIKI-TALK}, where memory synchronization induces more severe waiting bubbles in the pipeline. By shortening the critical-path memory-update stage, our optimization reduces the stall time behind memory synchronization and thus improves end-to-end pipeline efficiency. In particular, the critical-path memory-update time is reduced by \textbf{68.8\%}, \textbf{81.3\%}, and \textbf{88.3\%} on \texttt{WIKI}, \texttt{REDDIT}, and \texttt{WIKI-TALK}, respectively.

\begin{figure}[thbp]
\centering
\includegraphics[width=\linewidth]{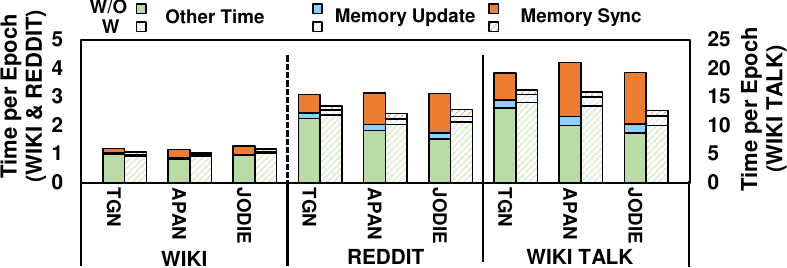}
\caption{Effect of critical-path optimization on multi-GPU pipeline execution. Comparison of pipeline execution with (\textsc{W}) and without (\textsc{W/O}) the optimization.}
\label{fig:exp-ab-critical}
\end{figure}

\textbf{Critical-path optimization for multi-GPU pipeline execution.}
We evaluate critical-path optimization under the same 8$\times$A100 setup and batch-size configuration as the main experiment. As shown in \cref{fig:exp-ab-critical}, we decompose epoch time into memory synchronization (\textit{Memory Sync}), critical-path memory update (\textit{Memory Update}), and remaining execution time (\textit{Other Time}). 
This optimization consistently improves pipeline efficiency, with average speedups of \textbf{1.11$\times$}, \textbf{1.22$\times$}, and \textbf{1.34$\times$} on \texttt{WIKI}, \texttt{REDDIT}, and \texttt{WIKI-TALK}, respectively. 
The benefit is larger on \texttt{REDDIT} and \texttt{WIKI-TALK}, where memory synchronization causes more severe pipeline bubbles. In particular, critical-path memory-update time drops by \textbf{68.8\%}, \textbf{81.3\%}, and \textbf{88.3\%} on \texttt{WIKI}, \texttt{REDDIT}, and \texttt{WIKI-TALK}, respectively. 
These results show that removing unnecessary operations from the pipeline critical path directly shortens dependency-induced stalls. The benefit is more pronounced for APAN and JODIE, whose complex memory updaters create a larger pipeline critical path. By decoupling non-urgent memory updates from the critical path and precomputing latency-sensitive metadata, DepBloom allows downstream GPUs to resolve dependencies earlier and resume execution sooner. The increasing gains on larger datasets further indicate that critical path optimization becomes more important as synchronization overhead grows.

\subsection{Hyperparameter Sensitivity Analysis} \label{subsec: eval_sens}

\begin{figure}[thbp]
\centering
\includegraphics[width=0.96\linewidth]{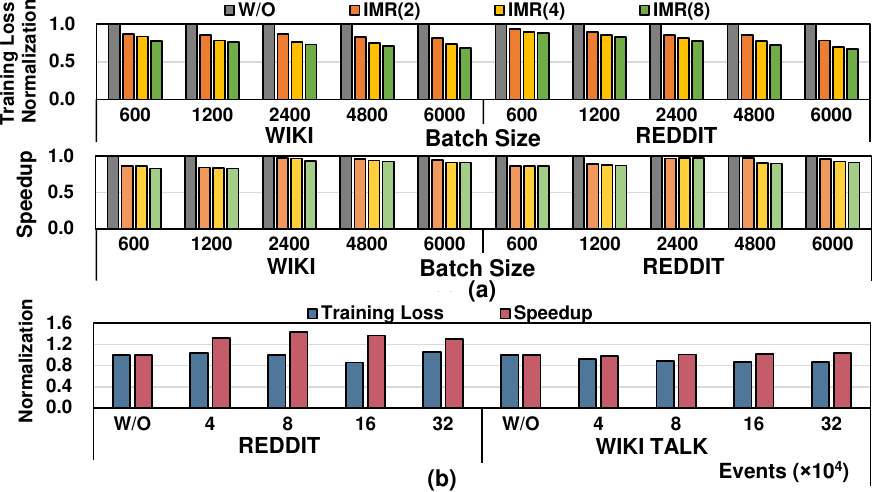}
\caption{
Sensitivity of intra-batch refinement and temporal bound. 
(a) Sensitivity to refinement granularity under different batch sizes on \texttt{WIKI} and \texttt{REDDIT}. \textit{Top:} Normalized training loss. \textit{Bottom:} Speedup.
(b) Sensitivity to the temporal bound on \texttt{REDDIT} and \texttt{WIKI-TALK}. Normalized training loss.
}
\label{fig:exp-h-refinement}
\end{figure}

\textbf{Sensitivity to refinement granularity.}
We vary the number of refinement chunks from \textsc{IMR(2)} to \textsc{IMR(4)} and \textsc{IMR(8)} under different batch sizes on \texttt{WIKI} and \texttt{REDDIT}. As shown in \cref{fig:exp-h-refinement}, finer-grained refinement consistently reduces training loss, especially at larger batch sizes, but introduces moderate overhead. Averaged across settings, the normalized loss decreases from \textbf{1.00} to \textbf{0.86}, \textbf{0.80}, and \textbf{0.76} with \textsc{IMR(2)}, \textsc{IMR(4)}, and \textsc{IMR(8)}, while speedup becomes \textbf{0.93}, \textbf{0.91}, and \textbf{0.90}, respectively. Overall, \textsc{IMR(4)} offers a good trade-off and is used as the default setting.

\textbf{Sensitivity to temporal bound.}
We vary the temporal bound in temporally bounded spatial partitioning and report results in \cref{fig:exp-h-refinement}(b). A moderate bound consistently provides a favorable accuracy-efficiency trade-off. On \texttt{REDDIT}, bound \textbf{8} achieves the highest speedup of \textbf{1.44}$\times$ with almost unchanged loss, while bound \textbf{16} further reduces loss to \textbf{0.86}$\times$ with a still high speedup of \textbf{1.37}$\times$. On \texttt{WIKI-TALK}, increasing the bound steadily improves both quality and efficiency in the evaluated range, with bound \textbf{32} achieving the best loss of \textbf{0.86}$\times$ and a speedup of \textbf{1.03}$\times$. Overall, these results suggest that a proper temporal bound can preserve useful temporal locality while exposing more parallelism.

\subsection{Overhead Analysis}

\begin{figure}[thbp]
\centering
\includegraphics[width=0.95\linewidth]{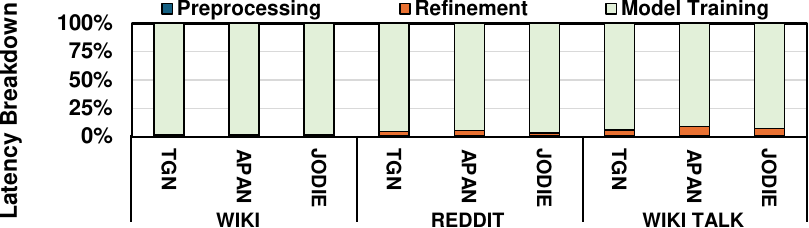}
\caption{Latency breakdown of \OurMethod{} across datasets and TGNN models, including preprocessing for temporally bounded spatial partitioning and extra computation from intra-batch memory refinement.}
\label{fig:exp-overhead}
\end{figure}

% \color{olive}
\OurMethod{} introduces two additional overhead sources: preprocessing for temporally bounded spatial partitioning and computation for intra-batch memory refinement.
As shown in \cref{fig:exp-overhead}, these overheads take only a small fraction of total training time across all evaluated datasets and models.
On average, preprocessing accounts for \textbf{0.61\%} of \OurMethod{}'s end-to-end training time, while memory refinement accounts for \textbf{3.78\%}; the remaining \textbf{95.61\%} is spent on normal training execution. Compared with the baseline training time, the added preprocessing and refinement costs are smaller, accounting for only \textbf{0.19\%} and \textbf{1.42\%} on average, respectively. 
The low preprocessing overhead arises because temporal window construction and topology-aware spatial partitioning run before training and are reused during execution. Refinement overhead is slightly higher because it runs during training. However, refinement enables larger batches while preserving short-range temporal information, amortizing its cost through higher parallelism and fewer synchronizations.

% \color{black}

% \OurMethod{} introduces two types of overhead. The first comes from the preprocessing for temporally bounded spatial partitioning, including temporal window construction and topology-aware spatial splitting. This step can be completed offline, and in practice accounts for only \textbf{1.02\%}, \textbf{0.64\%}, and \textbf{0.34\%} of end-to-end training time on \texttt{WIKI}, \texttt{REDDIT}, and \texttt{WIKI-TALK}, respectively. The second comes from intra-batch memory refinement, which introduces extra memory movement and slightly reduces training speed, as shown in \cref{fig:exp-ab-refinement} and the hyperparameter sensitivity study. However, since refinement substantially recovers the short-range temporal information lost in large-batch execution, it enables larger batches under the same accuracy target, allowing the added cost to be amortized by higher parallelism and fewer synchronizations.

\section{Related Work}

Although prior systems accelerate GNN training~\cite{yang2023betty,wang2021gnnadvisor,da2020distdgl,md2021distgnn}, they do not directly address multi-GPU TGNN training, whose core challenge comes from temporal dependencies across events. Earlier dynamic-graph studies largely focused on discrete-time dynamic graphs (DTDGs), which model graph evolution as snapshots~\cite{chen2023neutronstream,chakaravarthy2021efficient,wang2023pipad,rozemberczki2021pytorch,guan2022dynagraph,fu2023blad}. 
More recent TGNNs increasingly adopt continuous-time dynamic graphs (CTDGs), which model graph evolution as chronologically ordered event streams and thereby better preserve fine-grained temporal dynamics, but this also introduces event-level training dependencies that make snapshot-based DTDG optimizations difficult to transfer directly~\cite{trivedi2019dyrep,kumar2019predicting,xu2020inductive,rossi2020temporal,wang2021apan,zhou2022tgl,zhou2023disttgl,liu2025pipetgl,dai2025cascade,gao2024etc}.

Prior CTDG-focused work improves training efficiency through both algorithmic and systems-level techniques~\cite{zhou2022tgl,dai2025cascade,gao2024etc,sheng2024mspipe}. TGL~\cite{zhou2022tgl} provides a general framework for CTDG training, while ETC~\cite{gao2024etc} improves efficiency through dynamic batching and Cascade~\cite{dai2025cascade} enlarges batches by exploiting spatial and temporal independence.
PRISM~\cite{islam2026prism} mitigates intra-batch memory staleness through multi-version memory and refinement.
Other studies further reduce overheads by optimizing memory access, data movement, and runtime execution~\cite{liang2026tac,zhang2025effective,sheng2024mspipe,gao2024etc}. 
These methods are designed mainly for single-GPU execution. When extended to multi-GPU training, they do not address how batches should be organized across GPUs to expose parallelism, often leading to workload imbalance and low GPU utilization. 
In multi-GPU and distributed settings, GNNFlow~\cite{zhong2023gnnflow} improves graph storage and data movement, while DistTGL~\cite{zhou2023disttgl} and PipeTGL~\cite{liu2025pipetgl} optimize parallel execution over a fixed chronological batch sequence. 
Nevertheless, these methods either preserve event dependencies too conservatively or relax them too aggressively, and still struggle to jointly optimize training efficiency and model quality.
% \color{black}

% Prior CTDG-focused work improves training efficiency through both algorithmic and systems-level techniques~\cite{zhou2022tgl,dai2025cascade,gao2024etc,zhou2023disttgl,sheng2024mspipe,liu2025pipetgl,zhong2023gnnflow}. TGL~\cite{zhou2022tgl} provides a framework for CTDG training, and subsequent work enlarges batches through dynamic scheduling or dependency-aware partitioning~\cite{zhou2022tgl,dai2025cascade,gao2024etc,chen2023neutronstream}. ETC~\cite{gao2024etc} studies dynamic batching, while Cascade~\cite{dai2025cascade} enlarges batches based on spatial and temporal independence. Other studies improve efficiency by optimizing memory access, data movement, and runtime overheads~\cite{liang2026tac,zhang2025effective,sheng2024mspipe,gao2024etc}. These methods are designed mainly for single-GPU execution. When extended to multi-GPU training, they do not address how batches should be organized across GPUs to expose parallelism, often leading to load imbalance and low GPU utilization. In multi-GPU and distributed settings, GNNFlow~\cite{zhong2023gnnflow} improves graph storage and data movement, while DistTGL~\cite{zhou2023disttgl} and PipeTGL~\cite{liu2025pipetgl} optimize parallel execution over a fixed chronological batch sequence. Even so, they either strictly preserve event dependencies or ignore them, and still struggle to balance training efficiency and model quality.
\section{Conclusion}

% We present \OurMethod{}, a scalable multi-GPU training framework for TGNNs that improves efficiency by rethinking how temporal dependencies are enforced. Rather than synchronizing all dependencies eagerly, \OurMethod{} combines temporally bounded spatial partitioning, intra-batch memory refinement, and critical-path optimization to reduce unnecessary synchronization, recover missing short-range temporal information, and shorten memory-synchronization-induced stalls. Across representative TGNN models and dynamic graph datasets, \OurMethod{} achieves \textbf{2.63$\times$} average speedup and up to \textbf{5.16$\times$} speedup over state-of-the-art baselines while maintaining comparable or better model quality. We believe this result highlights a broader opportunity for scalable dynamic graph learning systems: improving performance by identifying which dependencies are truly critical, and relaxing the rest in a principled way.

We present \OurMethod{}, a scalable multi-GPU framework for TGNN training. Instead of synchronizing all dependencies across small batches, \OurMethod{} combines temporally bounded spatial partitioning, intra-batch memory refinement, and critical-path optimization to reduce unnecessary synchronization, recover missing short-range temporal information, and shorten memory-synchronization stalls. Across representative TGNN models and dynamic graph datasets, \OurMethod{} achieves \textbf{2.84$\times$} average speedup and up to \textbf{5.16$\times$} speedup over state-of-the-art baselines while maintaining comparable or better model quality. 
% These results highlight a broader opportunity for scalable dynamic graph learning: improving performance by identifying which dependencies are truly critical and relaxing the rest in a principled way.

%%
%% The next two lines define the bibliography style to be used, and
%% the bibliography file.
\bibliographystyle{ACM-Reference-Format}
\bibliography{sample-base}

%%
%% If your work has an appendix, this is the place to put it.
\appendix

\end{document}